\documentclass[times,twocolumn,final]{elsarticle}
\usepackage{comment}

\usepackage{geometry}
\newcommand{\verso}[1]{\def\@verso{#1}}

\newcommand{\KWD}{\vspace*{1em}}

\usepackage{framed,multirow}
\usepackage{amsmath,amssymb,amsfonts}
\usepackage{algorithmic}
\usepackage{caption}
\usepackage{hyperref}
\usepackage{booktabs}

\usepackage{algorithm}
\usepackage{algorithmic}
\usepackage{listings}
\begin{document}

\verso{Gaggion \MakeLowercase{\textit{et al.}}: Mask-HybridGNet}

\begin{frontmatter}



\title{Mask-HybridGNet: Graph-based segmentation with emergent anatomical correspondence from pixel-level supervision}

\author[1,2,3]{Nicolás Gaggion\corref{cor1}}
\author[4,5]{Maria J. Ledesma-Carbayo}
\author[6,7]{Stergios Christodoulidis}
\author[6,7]{Maria Vakalopoulou}
\author[3]{Enzo Ferrante}
\cortext[cor1]{Corresponding author: ngaggion@dc.uba.ar}

\address[1]{APOLO Biotech, Ciudad Autónoma de Buenos Aires, Argentina}
\address[2]{Departamento de Computación, Universidad de Buenos Aires, Ciudad Autónoma de Buenos Aires, Argentina}
\address[3]{Instituto de Ciencias de la Computación (ICC), CONICET-Universidad de Buenos Aires, Ciudad Autónoma de Buenos Aires, Argentina}
\address[4]{Biomedical Image Technologies, ETSI Telecomunicación, Universidad Politécnica de Madrid, Madrid, Spain}
\address[5] {Centro de Investigación Biomédica en Red de
Bioingeniera, Biomateriales y Nanomedicina (CIBER-BBN), Instituto de Salud Carlos III,Madrid, Spain}
\address[6]{IHU PRISM National Precision Medicine Center in Oncology,  Cancer Data Science Unit, CentraleSupélec, Gustave Roussy, INSERM, Université Paris-Saclay, Gif-sur-Yvette, France}
\address[7]{MICS, CentraleSup\'elec, Universit\'e Paris-Saclay, Gif-sur-Yvette, France}

\begin{abstract}
Graph-based medical image segmentation represents anatomical structures using boundary graphs, providing fixed-topology landmarks and inherent population-level correspondences. However, their clinical adoption has been hindered by a major requirement: training datasets with manually annotated landmarks that maintain point-to-point correspondences across patients rarely exist in practice. We introduce Mask-HybridGNet, a framework that trains graph-based models directly using standard pixel-wise masks, eliminating the need for manual landmark annotations. Our approach aligns variable-length ground truth boundaries with fixed-length landmark predictions by combining Chamfer distance supervision and edge-based regularization to ensure local smoothness and regular landmark distribution, further refined via differentiable rasterization. A significant emergent property of this framework is that predicted landmark positions become consistently associated with specific anatomical locations across patients without explicit correspondence supervision. This implicit atlas learning enables temporal tracking, cross-slice reconstruction, and morphological population analyses. Beyond direct segmentation, Mask-HybridGNet can extract correspondences from existing segmentation masks, allowing it to generate stable anatomical atlases from any high-quality pixel-based model. Experiments across chest radiography, cardiac ultrasound, cardiac MRI, and fetal imaging demonstrate that our model achieves competitive results against state-of-the-art pixel-based methods, while ensuring anatomical plausibility by enforcing boundary connectivity through a fixed graph adjacency matrix. This framework leverages the vast availability of standard segmentation masks to build structured models that maintain topological integrity and provide implicit correspondences. Our implementation is available at \url{https://huggingface.co/spaces/ngaggion/MaskHybridGNet}
\end{abstract}


\end{frontmatter}

\section{Introduction}

Medical image segmentation is fundamental to clinical workflows, enabling diagnosis, treatment planning, and patient monitoring \cite{ferrante2018weakly, cerrolaza2019computational}. Deep convolutional neural networks, particularly U-Net architectures \cite{ronneberger2015u}, and more recent Vision Transformer architectures \cite{valanarasu2021medical}, have achieved remarkable performance across various imaging modalities \cite{litjens2017survey, cxrsurvey}. However, these pixel-based approaches face a major limitation: they produce pixel-level segmentation masks and are trained using loss functions also defined at the pixel level (like cross-entropy or Dice coefficient \cite{milletari2016v}) without explicitly encoding anatomical knowledge or topological constraints. This can lead to anatomically implausible segmentations featuring discontinuities, holes, and irregular boundaries, particularly under challenging conditions such as poor image quality, partial occlusions, or domain shifts \cite{choudhary2020advancing, cohen2020limits, gaggion}.

Anatomical structures form contiguous regions with well-defined boundaries following consistent patterns across individuals. Graph-based methods leverage this property by representing anatomical boundaries as fixed-topology boundary graphs. This ensures topological correctness by construction, as the predefined connectivity of the graph prevents the formation of holes, isolated pixel clusters, or fragmented boundaries. \cite{boussaid2014discriminative, besbes2011landmark}. Recent graph-based methods, such as HybridGNet \cite{gaggion_miccai, gaggion}, demonstrate superior robustness to domain shifts and anatomical consistency compared to pixel-based approaches. Despite these advantages, graph-based methods remain confined to research settings due to a fundamental barrier: they require training datasets with manually annotated landmarks, maintaining point-to-point correspondence across patients. Such annotations require extensive expert time and are rarely found in practice.

\begin{figure*}[t!]
\centering
\includegraphics[width=\linewidth]{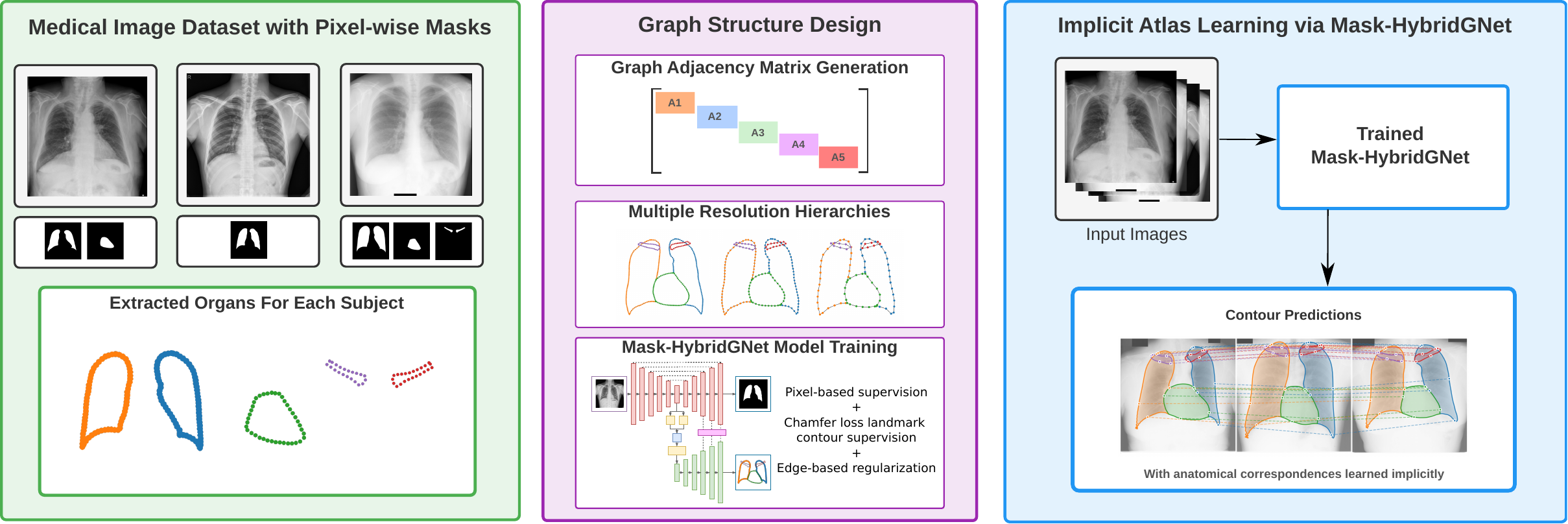}
\caption{\textbf{Framework overview.} Mask-HybridGNet enables training graph-based segmentation models with standard pixel-level supervision, without manual landmark annotations. (Left) Given a set of medical imaging datasets with potentially heterogeneous pixel-level masks, we first automatically extract variable-length contour pixels from the masks. (Middle) Graph structures are generated from dataset statistics, which determine fixed landmark counts and construct adjacency matrices. This allows for training our Mask-HybridGNet models. (Right) Our model produces fixed-topology boundary graphs where landmark indices implicitly represent consistent anatomical locations across patients, without explicit correspondence supervision.}
\label{fig:pipeline}
\end{figure*}

This paper presents Mask-HybridGNet, a framework that eliminates this barrier by enabling graph-based models to be trained directly on standard datasets containing only dense pixel-level annotations. Our approach is trained using conventional pixel-wise segmentation masks, but it produces structured boundary graphs through a systematic pipeline that discovers anatomical correspondences implicitly during training. Figure~\ref{fig:pipeline} illustrates how standard pixel-level masks can now be used to train sophisticated graph-based models without manual landmark annotation.

Our work makes three key contributions. First, we introduce a systematic pipeline that leverages standard medical imaging datasets to construct fixed-size graph topologies based on dataset statistics. This framework allows for the automatic derivation of adjacency matrices that can either represent organs as independent structures or as unified graphs with shared boundaries. Second, we develop a training strategy to address the mismatch between variable-length ground truth contours and fixed-length predictions. We use Chamfer distance supervision to align the predicted landmarks with the mask boundaries; however, because Chamfer distance is permutation invariant and does not enforce node order, it cannot produce a structured graph on its own. To resolve this, we incorporate edge-based regularization to enforce local smoothness and a regular distribution of landmarks. We further combine these with differentiable rasterization to improve the pixel-level quality of the segmentations. This process leads to the discovery of anatomical correspondences without explicit supervision. Third, we propose a novel dual-decoder architecture that optimizes feature sharing by routing image-to-graph skip-connections from an auxiliary pixel-wise decoder directly into the graph decoder.

By learning to map segmentation masks onto fixed-size contours, our model exhibits a remarkable emergent property: the ability to implicitly learn anatomical atlases. Without any supervision involving correspondence, the optimization process naturally causes each landmark position to become associated with specific anatomical locations that remain consistent across patients(\textit{i.e.,} the $i$-th landmark represents approximately the same anatomical point across all subjects). This allows the integration of point-based statistical priors \cite{cootes1992training, alven2019shape} into end-to-end deep learning architectures. The learned correspondences enable population-level shape analysis, statistical modeling of anatomical variations, and registration initialization, applications that traditionally required separate correspondence establishment \cite{zitova2003image, alven2016shape}.  

We demonstrate these capabilities through comprehensive experiments across diverse medical imaging applications. First, the Chest-Xray-landmark dataset \cite{gaggion}, containing both pixel masks and manual landmarks from four institutions (JSRT \cite{jsrt_shiraishi2000development}, Padchest \cite{bustos2020padchest}, Montgomery \cite{montgomeryset}, Shenzhen \cite{shenzhen}), enables direct comparison with landmark-supervised methods.  Second, we evaluate two different cardiac imaging datasets: CAMUS \cite{CAMUS} (echocardiography) demonstrates temporal correspondence tracking through cardiac cycles, while Sunnybrook \cite{sunnybrook} (MRI) illustrates consistent landmark positioning across different ventricular slices. Third, multi-dataset fetal head segmentation across three distinct protocols (HC18 \cite{hc18paper} for prenatal biometry and JNU-IFM \cite{jnu-ifm} and PSFHS \cite{psfhs} for intrapartum monitoring) validates cross-institutional generalization despite domain shifts. Finally, the application in PAX-Ray++ \cite{paxray} demonstrates scalability through simultaneous segmentation of 37 anatomical structures. 

We make our framework, trained models, and code publicly available to accelerate adoption and encourage further research in graph-based anatomical modeling.

\section{Related Work}

\subsection{Pixel-based Anatomical Segmentation}

Over the last decade, deep neural networks have revolutionized image segmentation. The initial fully-convolutional network architecture \cite{fcn} enabled end-to-end learning for pixel-wise segmentation. At the same time, the U-Net \cite{ronneberger2015u} model proposed an encoder-decoder architecture with skip connections to recover high-resolution details, which received strong adoption in the medical imaging community. Later, the nnU-Net framework \cite{isensee2021nnunet} became the state-of-the-art in pixel-level segmentation, adapting preprocessing, architecture, and training to the characteristics of each dataset. More recently, the Transformer architecture \cite{vaswani2017attention} has been adapted for medical image segmentation \cite{valanarasu2021medical}. These models leverage self-attention mechanisms to treat images as sequences of patches, enabling the capture of long-range dependencies that traditional convolutional layers may overlook. However, all these methods are trained with pixel-level loss functions, which, as shown by extensive literature \cite{imagemetrics, maier2022metrics}, have limited sensitivity to global shape and topology. This leads to artifacts such as fragmented structures and topological inconsistencies \cite{bohlender2021survey}, particularly in challenging imaging conditions, resulting in segmentation masks with low anatomical plausibility.

To overcome these limitations, researchers have developed various methods to add anatomical knowledge to deep learning segmentation models. Oktay et al. \cite{oktay2017anatomically} introduced anatomically constrained neural networks that add shape regularization terms derived from autoencoders trained on segmentation masks. Larrazabal et al. \cite{larrazabal2019anatomical,larrazabal2020post} showed that post-processing arbitrary anatomical masks with shape-constrained denoising autoencoders can recover anatomically plausible segmentations from noisy predictions. These approaches attempt to add anatomical constraints to pixel-level representations that do not inherently encode structural knowledge. They often require additional optimization procedures or computationally expensive post-processing steps. More importantly, they cannot guarantee topological correctness since the underlying pixel-level representation remains unchanged.

\subsection{Landmark and Graph-Based Anatomical Segmentation}

In parallel to the development of pixel-level segmentation models, since the early 1990s, structured representations of anatomical boundaries have been extensively investigated in medical image analysis. Point Distribution Models (PDMs) \cite{cootes1992training} introduced statistical modeling of landmark variations, which evolved into Active Shape Models (ASMs) \cite{cootes1992training, sozou1997non} and Active Appearance Models (AAMs) \cite{cootes1998active}. These methods combined shape priors with appearance models for deformable template matching. Multi-atlas segmentation methods \cite{alven2019shape, alven2016shape} extended template matching using multiple anatomical templates with registration algorithms \cite{zitova2003image}. These methods demonstrated the benefits of incorporating population-level anatomical knowledge into segmentation tasks. Earlier work combined classical shape modeling with deep learning. Milletari et al. \cite{milletari2017integrating} integrated CNNs with PCA-based shape representations, while Bhalodia et al. \cite{bhalodia2018deepssm,bhalodia2021deepssm} developed frameworks for automatic shape model construction. However, these approaches remain limited by PCA's linear modeling assumptions and require careful initialization and establishment of correspondence.

Recent approaches have leveraged geometric deep learning \cite{bronstein2017geometric} for anatomical segmentation by representing anatomical boundaries as fixed-topology boundary graphs. In our previous work \cite{gaggion_miccai, gaggion}, we introduced HybridGNet, which applies generative modeling to represent these anatomical shapes as boundary graphs. Building on the work of Ranjan et al. \cite{ranjan2018generating}, who demonstrated that mesh autoencoders using spectral convolutions \cite{bruna2013spectral, defferrard2016convolutional} can learn compact latent representations of 3D facial meshes, we extended HybridGNet to HybridVNet \cite{gaggion2025multi}. This demonstrated the versatility of the approach to model 3D organs as both volumetric and surface meshes. Both HybridVNet and HybridGNet utilize a variational encoder-decoder architecture, where a CNN-based encoder projects input images into a latent distribution. The graph decoder then samples from this distribution to infer landmark positions, constructing a graph representation of the anatomical boundaries. More recently, Bransby et al. \cite{Bransby_2023} addressed accuracy limitations of graph-based segmentation through joint dense-point learning. Their approach combines features from pixel-level and point-level training objectives, utilizing two independent networks: a UNet for dense segmentation and a HybridGNet for graph-based boundary prediction, with feature sharing between the UNet decoder and the graph encoder. 

The generative nature of these variational models provides significant advantages: learning a latent space of anatomical variations ensures anatomically plausible outputs and prevents the generation of impossible configurations. However, this modeling requires training data with anatomical correspondences. Each landmark must represent the same anatomical location across all training samples for the model to learn a meaningful latent space. 

\subsection{Contour-Based Methods with Differentiable Rendering}

A different class of methods employs a direct optimization approach to contour-based segmentation, rather than using generative modeling. These methods typically initialize contours and refine them iteratively, using various strategies to bridge the gap between contour representations and pixel-level supervision.
Curve-GCN \cite{curvegcn} treats contours as circular graphs, refining vertex positions via graph convolutions and a Point Matching Loss, later fine-tuned with a Differentiable Accuracy Loss using polygon triangulation. ACDRNet \cite{ACDRNet} predicts a global displacement field through an encoder–decoder, updating contours via sampled vertex displacements and adding active contour forces (balloon and curvature) with neural mesh rendering. BoundaryFormer \cite{boundaryformer} introduces a Transformer-based architecture that directly predicts polygons from masks using a differentiable rasterizer without triangulation, leveraging attention mechanisms to refine ellipsoidal polygons and achieving competitive results with Mask R-CNN. While these approaches enable training without explicit correspondences and achieve strong results on segmentation datasets, they optimize contour positions per image rather than capturing population-level shape priors, limiting their anatomical modeling capacities when compared to generative approaches. Finally, in the context of improving models trained on datasets with landmark correspondences, Bransby et al. \cite{Bransby_2023} proposed to combine the reconstruction error with a Hybrid Contour Distance loss that samples precomputed distance fields. This objective is formulated to accompany the standard Mean Squared Error (MSE) loss as a contour-aware term, specifically intended to prevent the penalization of predictions that align with the boundary even if they deviate from specific landmark positions.

\subsection{Positioning Our Contribution}

The contour-based segmentation methods discussed above fall into two paradigms: direct optimization, which trains on standard pixel masks but lacks generalizable shape modeling, and variational generative approaches, which capture population-level variations but traditionally require training datasets with explicit point-to-point anatomical correspondences. This requirement for manual landmark annotations leaves most clinical datasets, which typically provide only pixel-wise masks, inaccessible to these models.

Our work bridges this gap by training variational graph-based models directly with mask-derived supervision. We introduce a dual-decoder architecture where an auxiliary pixel-wise branch optimizes shared features for dense segmentation, which are then routed to the graph decoder. To supervise the graph branch without ordered landmarks, we utilize Chamfer distance for boundary alignment. Since Chamfer distance is permutation-invariant and lacks inherent node ordering, we incorporate edge-based regularization terms for elasticity, curvature, and uniform distribution. These constraints are deeply inspired by the classical active contour literature \cite{kass1988snakes}, providing the geometric guidance necessary to maintain local smoothness and structural integrity in a deep learning context. When combined with differentiable rasterization for pixel-level refinement, this process allows the model to implicitly discover consistent anatomical correspondences across patients. This emergent property transforms the framework from a standard segmentation tool into an automatic atlas generator, enabling population-level shape analysis and temporal tracking using only standard pixel-level annotations.

\section{Mask-HybridGNet: Learning Implicit Anatomical Correspondences for Graph-based Medical Image Segmentation}

\subsection{Problem Formulation}

Given a dataset $\mathcal{D} = \{(\mathbf{I}_n, \mathbf{M}_n)\}_{1 \leq n \leq N}$ of medical images $\mathbf{I}_n$ and pixel-wise segmentation masks $\mathbf{M}_n$, our objective is to learn a mapping from images to structured boundary graphs $\mathbf{G} = \langle V, \mathbf{A}, \mathbf{X} \rangle$. Here, $V$ is the set of nodes representing anatomical landmarks, $\mathbf{A}$ is the adjacency matrix encoding landmark connectivity (see Section \ref{sec:adjacencymat}), and $\mathbf{X} \in \mathbb{R}^{|V| \times 2}$ assigns spatial coordinates to the landmarks across all organs. To this end, we first process the pixel-wise segmentation masks $\mathbf{M}_n$ by extracting their variable-length contour pixels $\mathbf{C}_n^{(o)} = \{\mathbf{c}_1^{(o)}, \mathbf{c}_2^{(o)}, ..., \mathbf{c}_{L_n^{(o)}}^{(o)}\}$ for each organ $o \in \mathcal{O}$, where each $\mathbf{c}_i^{(o)} \in \mathbb{R}^2$ represents a 2D coordinate on the organ boundary, $\mathcal{O}$ denotes the set of anatomical structures, and $L_n^{(o)}$ denotes the contour length that varies across samples. These will be used as ground-truth for training our model. Throughout this section, superscripts in parentheses denote organ-specific indices to distinguish them from mathematical exponents. Dataset-specific processing details are provided in \ref{app:config}.

While the ground-truth contours extracted from the organ masks have variable length, our model will learn to produce atlases that are represented as a fixed-size graph. To this end, to enable consistent graph processing, we define fixed-size landmark representations. For each organ $o \in \mathcal{O}$, we determine the number of landmarks as:

\begin{equation}
N_{1}^{(o)} = \max\left(\left\lfloor\bar{L}^{(o)} \cdot s\right\rfloor, N_{\min}\right)
\end{equation}

\noindent where $\bar{L}^{(o)} = \frac{1}{|\mathcal{T}|} \sum_{t \in \mathcal{T}} L_t^{(o)}$ represents the average contour length for organ $o$ over the training set $\mathcal{T}$, $s$ is a scale factor, and $N_{\min}$ is the minimum number of landmarks to ensure minimum representation quality.

Our model outputs boundary graph predictions at multiple resolution levels $r \in \{1, 2, ..., R\}$ to capture anatomical variations across different spatial scales. Following the hierarchical design of graph convolutional autoencoders \cite{ranjan2018generating, gaggion}, this multi-scale formulation enables the decoder to aggregate features over increasing receptive fields. By processing coarser representations at lower resolutions, the model captures global anatomical shape constraints, while higher resolutions refine fine-grained boundary details using high-frequency image features from the image-to-graph skip connections. For each organ $o$ and resolution level $r$, we define the landmark count as:

\begin{equation}
N_{r}^{(o)} = \lfloor N_{1}^{(o)} / 2^{r-1} \rfloor
\end{equation}

\noindent where $r=1$ is the full resolution and higher values correspond to progressively coarser graphs.

\subsection{Adjacency Matrix Construction}
\label{sec:adjacencymat}

Modeling anatomy as boundary graphs requires balancing internal topology with inter-organ relationships. While some organs can be treated as independent closed contours, others share boundaries that demand joint modeling. To capture both scenarios, we define two adjacency matrix constructions: one for independent organs, ensuring anatomical plausibility in topological terms, and another for unified multi-organ graphs, encoding shared boundaries and spatial constraints.\\

\noindent \textbf{Independent Graph Representation:} For anatomical structures without shared boundaries, each organ is modeled as an independent circular graph. The adjacency matrix follows a block-diagonal structure:

\begin{equation}
\label{eq:matrix}
\mathbf{A} = 
\begin{bmatrix}
\mathbf{A}^{1} & \mathbf{0} & \cdots & \mathbf{0} \\
\mathbf{0} & \mathbf{A}^{2} & \cdots & \mathbf{0} \\
\vdots & \vdots & \ddots & \vdots \\
\mathbf{0} & \mathbf{0} & \cdots & \mathbf{A}^{|\mathcal{O}|}
\end{bmatrix}
\end{equation}

\noindent where each organ-specific adjacency matrix $\mathbf{A}^{(o)} \in \{0,1\}^{N_{1}^{(o)} \times N_{1}^{(o)}}$ encodes circular connectivity:

\begin{equation}
\mathbf{A}^{(o)}_{i,j} = 
\begin{cases}
1 & \text{if } j = (i-1) \bmod N_{1}^{(o)} \text{ or } j = (i+1) \bmod N_{1}^{(o)} \\
0 & \text{otherwise.}
\end{cases}
\end{equation}

This ensures each landmark has exactly two neighbors, maintaining closed contour topology while keeping organs completely isolated.

For hierarchical processing, we construct a hierarchy of boundary graphs with decreasing landmark counts. We define organ-specific downsampling and upsampling matrices to transition features between these scales. The downsampling matrices $\mathbf{D}^{(o,r \rightarrow r+1)}$ implement graph pooling by combining adjacent nodes:

\begin{equation}
\mathbf{D}^{(o,r \rightarrow r+1)}_{i,j} = 
\begin{cases}
\frac{1}{2} & \text{if } j = 2i \text{ or } j = (2i+1) \bmod N_{\text{r}}^{(o)} \\
0 & \text{otherwise.}
\end{cases}
\end{equation}

The upsampling matrices $\mathbf{U}^{(o,r+1 \rightarrow r)}$ distribute features from coarser to finer representations:

\begin{equation}
\mathbf{U}^{(o,r+1 \rightarrow r)}_{i,j} = 
\begin{cases}
1 & \text{if } i = \lfloor j/2 \rfloor \\
0 & \text{otherwise.}
\end{cases}
\end{equation}

\noindent \textbf{Unified Graph Representation:} While independent graphs capture individual organ topology, many anatomical structures benefit from joint modeling via a single, integrated boundary graph. Medical images often include structures with shared boundaries, such as cardiac chambers where endocardium and myocardium meet. The independent block-diagonal structure (see Eq. \ref{eq:matrix}) treats such contacting structures as separate entities, potentially missing valuable spatial constraints. This leads to a duplication of information, where the same anatomical interface is represented independently in multiple separate graphs without recognition of their spatial relationship. For structures with defined adjacencies, we construct a unified boundary graph where specific nodes belong to multiple organs, explicitly encoding shared boundaries within the adjacency matrix. For organs that may or may not be in contact, independent boundary graphs remain appropriate.

For anatomical structures with shared boundaries, we construct a unified graph where nodes may belong to multiple organs. Each node $v \in V$ has an associated membership set $\mathcal{M}_v \subseteq \mathcal{O}$ indicating organ affiliation. The shared boundary regions between organs $i, j \in \mathcal{O}$ are defined as:

\begin{equation}
\mathcal{B}_{i,j} = \{v \in V : i \in \mathcal{M}_v \land j \in \mathcal{M}_v\}
\end{equation}

The resulting adjacency matrix $\mathbf{A} \in \{0,1\}^{|V| \times |V|}$ captures both intra-organ connectivity and inter-organ boundary relationships, enabling joint modeling of anatomically connected structures. Detailed construction algorithms for these unified boundary graphs and the identification of shared interfaces are provided in \ref{app:unified_construction}.

For multi-resolution processing of unified boundary graphs, we implement topologically-aware coarsening to ensure that structural interfaces are not lost during pooling. In this context, nodes are classified by their degree: nodes with exactly two neighbors (simple contour segments) are eligible for downsampling, whereas junction points (nodes with more than two neighbors where multiple organs meet) are preserved across all resolution levels. This selective downsampling ensures that the shared anatomical boundaries and the global connectivity of the unified boundary graph remain structurally consistent across the hierarchy. The specific construction of the corresponding downsampling and upsampling operators is detailed in \ref{app:topological_processing}.

\begin{figure*}[t]
    \centering
    \includegraphics[width=\linewidth]{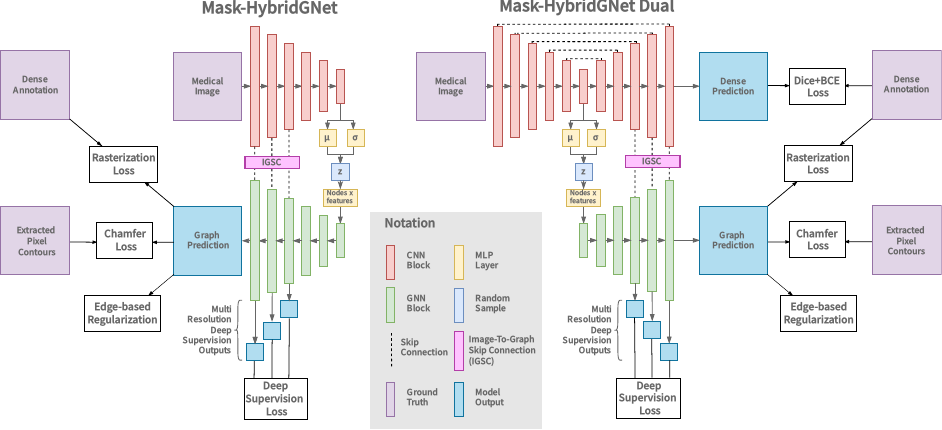}
    \caption{\textbf{Architectural overview of the Mask-HybridGNet framework.} 
     (Left) The standard Mask-HybridGNet architecture couples a CNN encoder with a graph decoder via Image-to-Graph Skip Connections (IGSC). (Right) The Dual variant introduces an auxiliary CNN decoder following a U-Net topology. In this setup, the IGSC layers are redirected to sample feature maps from the auxiliary decoder instead of the encoder. Both variants share the same graph decoder, utilize a variational bottleneck for shape modeling, and are trained using a combination of Chamfer distance, edge-based regularization, and differentiable rasterization. The Dual model additionally incorporates a pixel-wise loss for the auxiliary decoder branch.
    }
    \label{fig:architecture}
\end{figure*}

\subsection{Graph Neural Network Architecture}
\label{sec:gnn_arch}

The Mask-HybridGNet framework builds upon the HybridGNet architecture \cite{gaggion_miccai, gaggion}, which integrates convolutional feature extraction with graph-based geometric decoding. Our implementation utilizes a unified backbone consisting of a CNN encoder with a variational bottleneck, branching into different decoding strategies that optimize how spatial information is propagated to the graph representation, as illustrated in Figure~\ref{fig:architecture}. \\

\noindent \textbf{Shared Variational Backbone:} The encoder maps an input image $\mathbf{I}$ to a latent distribution through a series of convolutional blocks. The final spatial feature map is passed through parallel fully connected layers to derive the distribution parameters $\boldsymbol{\mu}$ and $\log\boldsymbol{\sigma}^2$. A latent vector $\mathbf{z}$ is then sampled using the reparameterization trick:

\begin{equation}
\mathbf{z} = \boldsymbol{\mu} + \boldsymbol{\sigma} \odot \boldsymbol{\epsilon}, \quad \boldsymbol{\epsilon} \sim \mathcal{N}(\mathbf{0}, \mathbf{1})
\end{equation}

This latent vector $\mathbf{z}$ serves as a compressed representation of the global anatomical shape. To initialize the graph decoding process, $\mathbf{z}$ is passed through a fully connected layer that projects the global vector into a structured feature matrix at the coarsest resolution level $N^{(o,R)}$, providing the initial vertex-wise features used as input for the first graph convolutional layer. \\

\noindent \textbf{Decoding Configurations:} We evaluate two architectural variants, Mask-HybridGNet and Mask-HybridGNet Dual, which differ in the signal flow of the Image-to-Graph Skip Connections (IGSC) \cite{gaggion}.

In the Mask-HybridGNet configuration, the graph decoder utilizes IGSC to sample multi-scale spatial features directly from the CNN encoder blocks. This setup allows the graph branch to refine landmark positions using hierarchical features extracted during the encoding process.

The Mask-HybridGNet Dual configuration introduces an auxiliary CNN decoder following a standard U-Net-like architecture. This branch begins at the spatial bottleneck and progressively upsamples the representation, integrating encoder features via standard skip connections to generate a dense mask prediction $\mathbf{S}$. By supervising $\mathbf{S}$ with pixel-level losses, the resulting decoder feature maps are explicitly optimized for boundary localization. In this Dual variant, the IGSC layers are redirected to sample from these refined decoder feature maps rather than the encoder feature maps. This enables the graph decoder to leverage spatial representations already conditioned for the segmentation task to predict the multi-resolution landmark positions $\mathbf{P}$. While the CNN branch is required to guide this optimization during training, only the structured graph is retained for the final clinical output. \\

\noindent \textbf{Graph Decoder Layers:} The graph decoder branch in both configurations employs Chebyshev spectral graph convolution layers \cite{defferrard2016convolutional}. These layers aggregate the sampled spatial features within the local neighborhood of each node according to the predefined adjacency matrix $\mathbf{A}$. \\

\noindent \textbf{Differentiable Rasterization:} To bridge the gap between point-based and pixel-based representations, we follow \cite{boundaryformer} and implement a differentiable SoftPolygon rasterizer that converts predicted landmark coordinates to pixel-wise masks:

\begin{equation}
\mathbf{M} = \text{SoftPolygon}(\mathbf{P}, H, W, \sigma)
\end{equation}

This rasterizer takes predicted landmark points $\mathbf{P}$, desired output dimensions $H \times W$, and a smoothness parameter $\sigma$. It produces a differentiable mask where pixels inside the polygon defined by the contour get assigned positive values close to 1, while those outside get assigned values close to 0. The smoothness parameter controls the transition region around the contour, with smaller values creating sharper boundaries. This enables end-to-end training with both contour-based and pixel-wise losses.

\subsection{Loss Function: Data Terms}
\label{sec:training}

One of the main challenges faced when training Mask-HybridGNet is learning from variable-length ground truth contours while producing fixed-length predictions. To overcome this limitation, our approach combines Chamfer distance supervision with complementary regularization terms and progressive optimization strategies.\\

\noindent \textbf{Chamfer Distance Loss.} The primary supervision signal uses Chamfer distance to handle the asymmetry between variable-length ground truth contours and fixed-length model predictions. We calculate this distance independently for each anatomical structure to ensure organ specific alignment. Given a set of predicted landmarks $\mathbf{P}$ and a set of ground truth boundary pixels $\mathbf{G}$, the Chamfer distance is defined as:

\begin{equation}
d_{CD}(\mathbf{P}, \mathbf{G}) = \frac{1}{|\mathbf{P}|} \sum_{\mathbf{p} \in \mathbf{P}} \min_{\mathbf{g} \in \mathbf{G}} \|\mathbf{p} - \mathbf{g}\|^2 + \frac{1}{|\mathbf{G}|} \sum_{\mathbf{g} \in \mathbf{G}} \min_{\mathbf{p} \in \mathbf{P}} \|\mathbf{g} - \mathbf{p}\|^2.
\end{equation}

To handle heterogeneous datasets where certain organs may not be annotated in every image, the total Chamfer loss is computed as the sum of distances between corresponding pairs of predicted and ground truth sets for each organ $o$ present in the sample $\mathcal{O}$:

\begin{equation}
\mathcal{L}_{\text{chamfer}} = \sum_{o \in \mathcal{O}} d_{CD}(\mathbf{P}_o, \mathbf{G}_o).
\end{equation}

\noindent where $\mathbf{P}_o$ and $\mathbf{G}_o$ represent the predicted landmarks and ground truth pixels for organ $o$, respectively. \\

\noindent \textbf{Pixel-Level Loss.} We apply standard pixel-level segmentation losses to both the rasterized contour predictions and the dense output of the auxiliary CNN decoder. For a given organ $o \in \mathcal{O}$, we compute the Dice loss $\mathcal{L}_{\text{dice}}^{(o)}$ and Binary Cross Entropy loss $\mathcal{L}_{\text{bce}}^{(o)}$ by comparing the predicted pixel-wise mask against the ground truth.

For single-decoder architectures, these losses are computed solely on the rasterized contour predictions. In the dual-decoder setup, the loss is applied to both the rasterized masks and the direct CNN segmentation output $\mathbf{S}$. The total pixel-level loss for a sample is averaged over the set of annotated organs:

\begin{equation}
\mathcal{L}_{\text{pixel}} = \frac{1}{|\mathcal{O}|} \sum_{o \in \mathcal{O}} \left[ \mathcal{L}_{\text{dice}}^{(o)} + \mathcal{L}_{\text{bce}}^{(o)} \right].
\end{equation}

Note that rasterization is performed only at the highest resolution to maintain computational efficiency during training and backpropagation.

\subsection{Loss Function: Regularization Terms}

\noindent \textbf{Variational Regularization}. The variational encoder requires KL divergence regularization to maintain a well-structured latent space by penalizing the deviation of the learned distribution from a unit Gaussian prior:

\begin{equation}
\mathcal{L}_{\text{kld}} = -\frac{1}{2} \sum (1 + \log(\boldsymbol{\sigma}^2) - \boldsymbol{\mu}^2 - \boldsymbol{\sigma}^2)
\end{equation}

\noindent where the summation is performed over the dimensions of the latent vectors $\boldsymbol{\mu}$ and $\boldsymbol{\sigma}$.\\

\noindent \textbf{Edge-based Contour Regularization.} We enforce anatomical plausibility through regularization terms that operate on the graph structure. While the model produces a set of $N$ nodes with an associated adjacency matrix $\mathbf{A}$, direct computation using sparse operations is computationally inefficient. To optimize this, we utilize a tensor-based indexing scheme with a precomputed edge connectivity matrix $\mathbf{E} \in \mathbb{N}^{O \times M \times 2}$. This matrix organizes the $M$ edges of the $O$ organs into a dense format; organs with fewer than $M$ edges are zero-padded to maintain a consistent tensor shape for efficient GPU-accelerated processing.

For each organ $o$, the edge set $\mathcal{E}_o$ is defined by the pairs of node indices $(i, j)$ stored in $\mathbf{E}$:
\begin{equation}
\mathcal{E}_o = \{ (i,j) : \mathbf{E}_{o,k} = (i,j), \quad k=1, \dots, M \}.
\end{equation}

By indexing into the predicted node set $\mathbf{P} \in \mathbb{R}^{N \times 2}$, we compute the edge vector $\mathbf{e}_{ij}$ for every connected pair of landmarks:

\begin{equation}
\mathbf{e}_{ij} = \mathbf{p}_i - \mathbf{p}_j, \quad \forall (i,j) \in \mathcal{E}_o.
\end{equation}

Deeply inspired by classical active contour literature, these vectors $\mathbf{e}_{ij}$ act as the internal energy of the contours through three complementary regularization terms: \\

\paragraph{I) Uniform Edge Length Loss:} Encourages landmarks to be regularly distributed by penalizing deviations from the mean edge length:

\begin{equation}
\mathcal{L}_{\text{uniform}} = \frac{1}{O} \sum_{o=1}^{O} w_o \cdot \frac{1}{|\mathcal{E}_o|} \sum_{(i,j) \in \mathcal{E}_o} \left(\frac{\|\mathbf{e}_{ij}\|_2 - \bar{e}_o}{\bar{e}_o}\right)^2.
\end{equation}

\noindent where $\bar{e}_o = \frac{1}{|\mathcal{E}_o|} \sum_{(i,j) \in \mathcal{E}_o} \|\mathbf{e}_{ij}\|_2$ is the mean edge length for organ $o$ and $w_o$ are organ-adaptive weights (defined below).

\paragraph{II) Elasticity Loss:} Acts as internal tension by penalizing long edges, encouraging compact contours:

\begin{equation}
\mathcal{L}_{\text{elastic}} = \frac{1}{O} \sum_{o=1}^{O} w_o \cdot \frac{1}{|\mathcal{E}_o|} \sum_{(i,j) \in \mathcal{E}_o} \|\mathbf{e}_{ij}\|_2^2.
\end{equation}

\paragraph{III) Curvature Loss:} Promotes local smoothness by penalizing abrupt direction changes between consecutive edges $(i,j)$ and $(j,k)$ within the same organ:
\begin{equation}
\mathcal{L}_{\text{curvature}} = \frac{1}{O} \sum_{o=1}^{O} w_o \cdot \frac{1}{|\mathcal{C}_o|} \sum_{\substack{(i,j),(j,k) \in \mathcal{E}_o \\ i \neq k}} \|\mathbf{e}_{ij} - \mathbf{e}_{jk}\|_2^2.
\end{equation}

\noindent \textbf{Organ-Adaptive Weighting:} To address the inherent size imbalance between organs in the regularization terms, we introduce adaptive weights based on the contour perimeter:

\begin{equation}
w_o = \frac{P_o}{\max_{o' \in \{1,...,O\}} P_{o'}}, \quad \text{where} \quad P_o = \sum_{(i,j) \in \mathcal{E}_o} \|\mathbf{e}_{ij}\|_2.
\end{equation}

This weighting scheme ensures that larger organs contribute proportionally more to the total loss, preventing optimization bias towards smaller structures which showed to lead to over-regularized shapes.\\

The total edge-based regularization loss combines all three terms:

\begin{equation}
\mathcal{L}_{\text{edge}} = \alpha \mathcal{L}_{\text{uniform}} + \beta \mathcal{L}_{\text{elastic}} + \gamma \mathcal{L}_{\text{curvature}},
\end{equation}

\noindent where $\alpha$, $\beta$, and $\gamma$ are hyperparameters controlling the relative importance of each term. Unlike supervised losses, these regularization terms are computed for all organs regardless of ground truth availability, ensuring anatomical consistency across the complete organ set even when training on heterogeneous datasets where different samples contain different anatomical structures.

\subsection{Training Details and Optimization Strategy}

The complete loss function combines all terms with scheduled weighting:

\begin{equation}
\mathcal{L} = \lambda_c \mathcal{L}_{\text{chamfer}} + \lambda_p \mathcal{L}_{\text{pixel}} + \lambda_k \mathcal{L}_{\text{kl}} + \lambda_e \mathcal{L}_{\text{edge}},
\end{equation}

\noindent where $\mathcal{L}_{\text{edge}} = \alpha \mathcal{L}_{\text{uniform}} + \beta \mathcal{L}_{\text{elastic}} + \gamma \mathcal{L}_{\text{curvature}}$ as defined above.

We employ a progressive training strategy with scheduled loss weighting to ensure stable convergence. Training begins with batch size 1 for improved stability, then increases to the target size. Loss weights ($\lambda_c = 10.0$, $\lambda_p = 1.0$, $\lambda_k$ initially $10^{-6}$, $\alpha$ initially $10^{-6}$, $\beta = 300.0$, $\gamma = 250.0$) were determined through grid search across chest X-ray, cardiac ultrasound, cardiac MRI, and fetal imaging datasets. This search utilized the independent validation sets of each dataset to ensure the configuration generalizes across modalities. Loss weights follow specific schedules:

\begin{itemize}
\item \textbf{Regularization annealing}: Elasticity and curvature terms ($\beta$ and $\gamma$) decrease exponentially during training to allow initial shape learning without overly restrictive constraints.

\item \textbf{KL scheduling}: The KL weight ($\lambda_k$) increases exponentially from $10^{-6}$ to $10^{-3}$ to balance reconstruction quality and latent space regularization.

\item \textbf{Uniform activation}: The uniform edge length term ($\alpha$) activates progressively from $10^{-6}$ to $1.0$ during the first third of training using log-annealing, then remains constant.

\item \textbf{Primary supervision}: Chamfer distance ($\lambda_c$) and pixel-level ($\lambda_p$) weights remain constant throughout training to provide consistent supervision signals.
\end{itemize}

This scheduling approach prevents competing objectives during early training stages while ensuring proper geometric regularization as the model converges to anatomically plausible solutions.\\

\noindent \textbf{Implementation details.} We employ Adam optimization with learning rate $\eta = 10^{-4}$ for approximately 600,000 iterations, selecting the best model based on validation performance without early stopping. Input images are padded to a square, then reshaped to 512$\times$512 resolution for consistent processing. Data augmentation is configured per-dataset to respect anatomical constraints; for instance, chest X-ray augmentation excludes horizontal flipping to preserve left-right lung distinction, while allowing vertical flipping, limited rotation, and excluding transposition. The framework is implemented in PyTorch \cite{pytorch} with PyTorch Geometric \cite{pytorch-geometric} for graph operations, PyTorch3D \cite{pytorch3d} for Chamfer distance computation, and BoundaryFormer's \cite{boundaryformer} SoftPolygon for differentiable rasterization. Training requires 12-24 hours per dataset on a single NVIDIA RTX 3090 (24GB VRAM), with time scaling proportionally to the number of organs due to rasterization loss backpropagation overhead. Inference runs at approximately 50ms per image using 4GB VRAM for single-batch processing. All experiments were conducted on an Intel Core i9-14900KF workstation with NVIDIA RTX 3090 GPU; minimum hardware requirements are 12GB VRAM for training. Source code available at \url{https://github.com/ngaggion/MaskHybridGNet}. Online implementation available at \url{https://huggingface.co/spaces/ngaggion/MaskHybridGNet}.

\section{Experiments}

\subsection{Training Landmark-based Models with Pixel-level Masks for Chest X-Ray Segmentation}

\begin{figure}[t!]
\centering
\includegraphics[width=\linewidth]{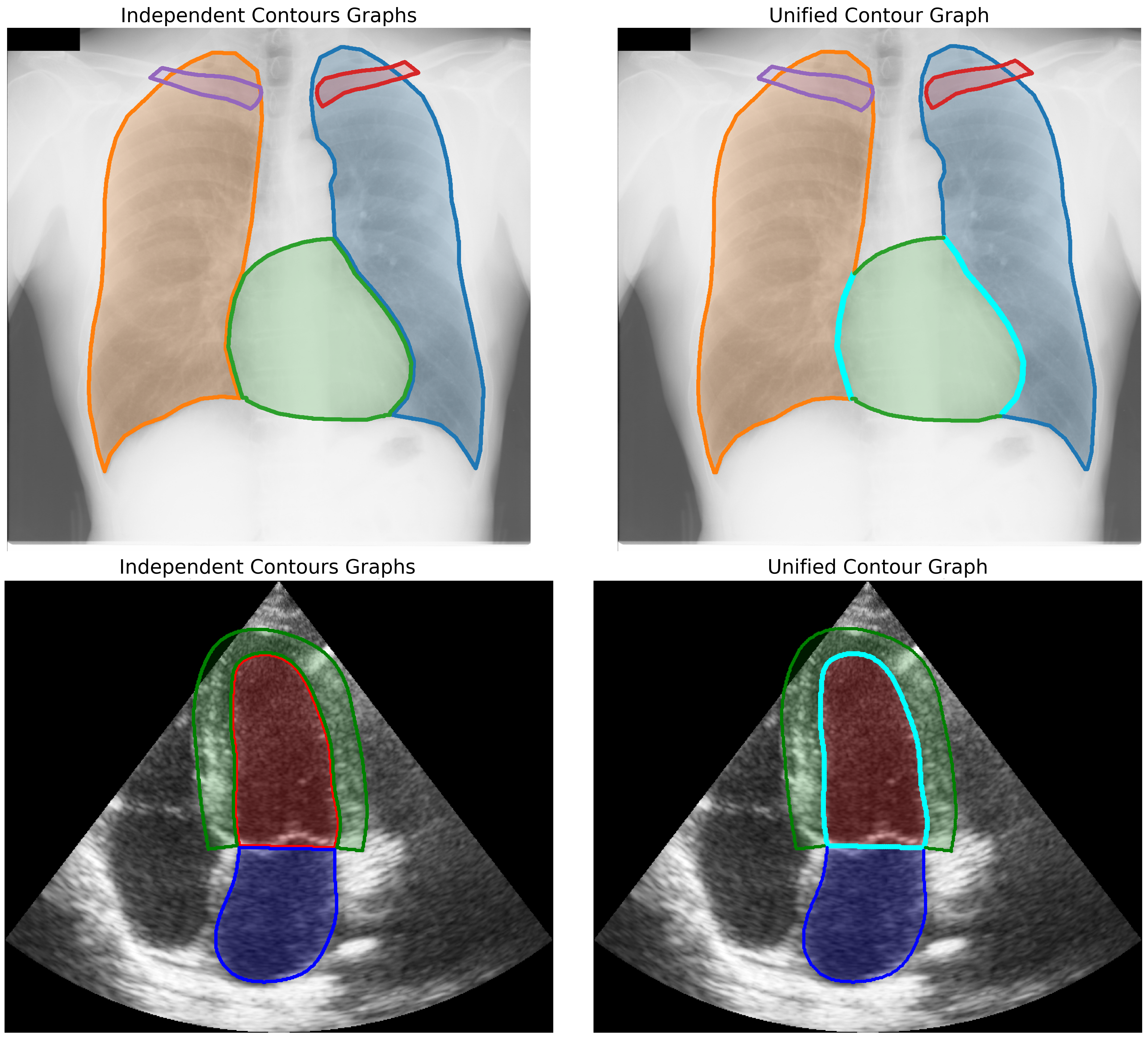}
\caption{\textbf{Graph representations for chest X-ray and echocardiograph anatomical structures.} The figure illustrates the two graph representation strategies employed in our framework. \textit{Left}: Independent graph representations treat each anatomical structure separately with circular graphs. For chest X-ray: left lung (blue), right lung (green), heart (orange), left clavicle (cyan), and right clavicle (purple). For echocardiography: left ventricular endocardium (red), left ventricular epicardium (green), and left atrium (blue). \textit{Right}: Unified graph representations model share anatomical boundaries, where nodes can belong to multiple organs. Cyan contour segments indicate shared interfaces between adjacent structures, enabling joint modeling of anatomically connected regions.}
\label{fig:organ_contours}
\end{figure}

\begin{table*}[t!]
\centering
\caption{Dice Coefficient Scores on the test-set of the Chest-xray-landmark database (mean ± std).}
\label{tab:chestxraycomparison}
\resizebox{\linewidth}{!}{
\begin{tabular}{ll cccc cc c}
\toprule
& & \multicolumn{4}{c}{\textbf{Lungs}} & \multicolumn{2}{c}{\textbf{Heart}} & \textbf{Clavicles} \\ 
\cmidrule(lr){3-6} \cmidrule(lr){7-8} \cmidrule(lr){9-9}
\textbf{Model} & \textbf{Graph Type} & \textbf{Shenzhen} & \textbf{Montgomery} & \textbf{Padchest} & \textbf{JSRT} & \textbf{Padchest} & \textbf{JSRT} & \textbf{JSRT} \\ 
\midrule
Baseline HybridGNet (MSE) & Independent & 0.963 ± 0.015 & 0.964 ± 0.031 & 0.956 ± 0.023 & 0.973 ± 0.010 & 0.939 ± 0.022 & 0.936 ± 0.030 & 0.816 ± 0.068 \\ 
nnUNet & - & 0.969 ± 0.014 & 0.982 ± 0.009 & 0.965 ± 0.017 & 0.979 ± 0.009 & 0.947 ± 0.016 & 0.953 ± 0.025 & 0.951 ± 0.018 \\ 
\midrule
Mask-HybridGNet & Independent & 0.961 ± 0.015 & 0.968 ± 0.018 & 0.953 ± 0.016 & 0.967 ± 0.014 & 0.925 ± 0.038 & 0.928 ± 0.038 & 0.801 ± 0.067 \\ 
Mask-HybridGNet Dual & Independent & 0.965 ± 0.014 & 0.974 ± 0.019 & 0.956 ± 0.022 & 0.975 ± 0.011 & 0.939 ± 0.019 & 0.941 ± 0.027 & 0.860 ± 0.077 \\ 
Mask-HybridGNet & Unified & 0.963 ± 0.014 & 0.974 ± 0.015 & 0.954 ± 0.017 & 0.972 ± 0.011 & 0.940 ± 0.022 & 0.935 ± 0.034 & 0.836 ± 0.068 \\ 
Mask-HybridGNet Dual & Unified & \textbf{0.965 ± 0.013} & \textbf{0.975 ± 0.018} & \textbf{0.955 ± 0.020} & \textbf{0.975 ± 0.011} & \textbf{0.942 ± 0.017} & \textbf{0.940 ± 0.027} & \textbf{0.862 ± 0.078} \\ 
\bottomrule
\end{tabular}}
\end{table*}

\begin{figure*}[ht!]
\centering
\includegraphics[width=0.9\linewidth]{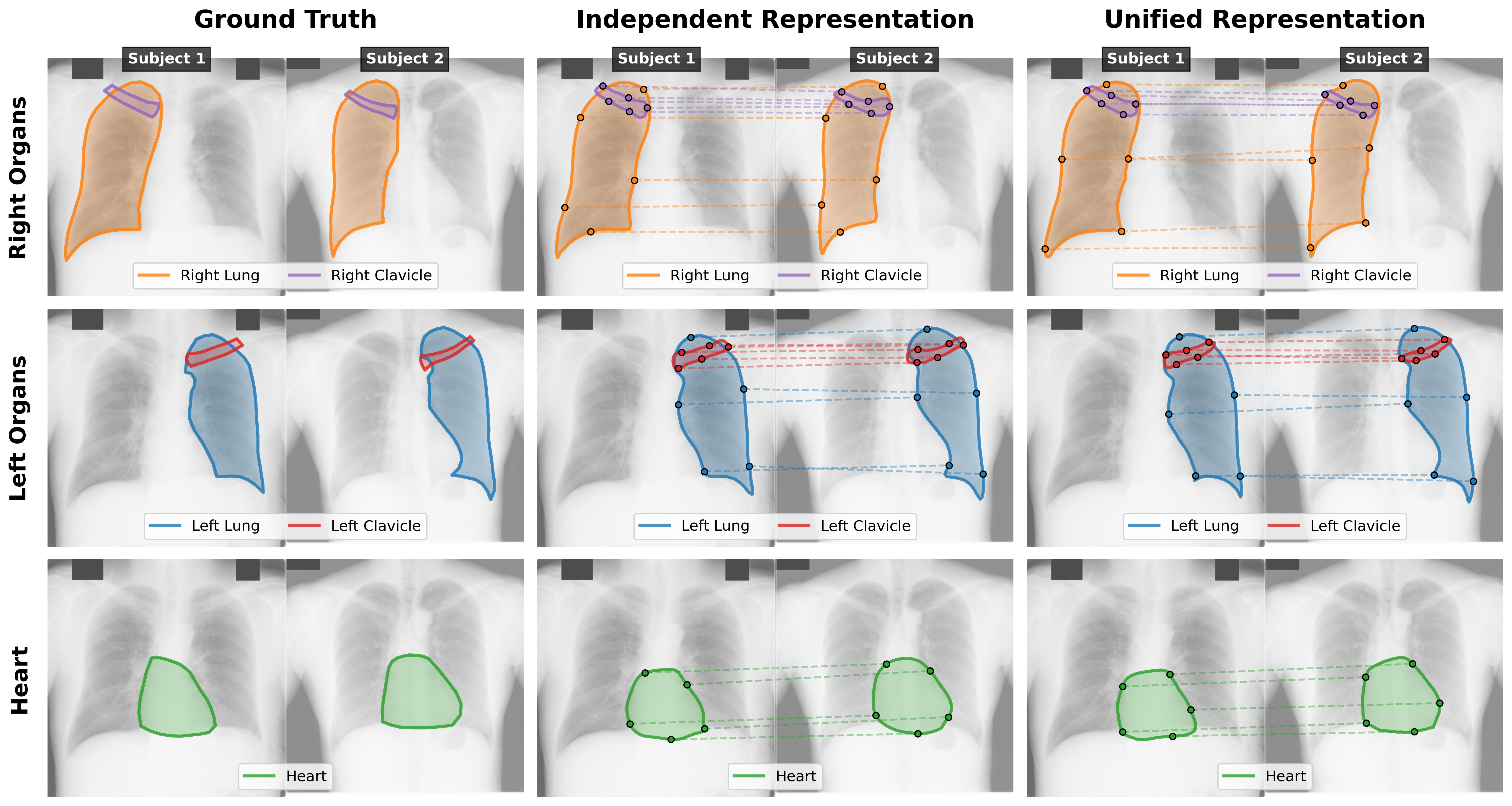}
\caption{\textbf{Emergent anatomical correspondences learned through mask supervision.} The figure demonstrates that our framework successfully learns anatomically meaningful landmark correspondences without explicit point-level supervision. The left column shows the ground-truth contours, middle column results from independent graph representations, while the right column show unified graph representations. Each row displays a subset of organs for two different subjects, revealing consistent anatomical locations on a subset of equidistant nodes.}
\label{fig:correspondences}
\end{figure*}

Our first experiment aims to validate the proposed framework's ability to train graph-based segmentation models using standard pixel-wise segmentation masks instead of manually annotated landmarks, which are easier to obtain and widely available. We compare our Mask-HybridGNet against the baseline HybridGNet \cite{gaggion,gaggion_isbi}, which was trained with landmark supervision using mean squared error (MSE) loss on manually annotated landmarks with point-to-point correspondences across patients.

We conduct this validation on the Chest-xray-landmark dataset, which contains landmark annotations and corresponding pixel-level masks for 911 images from 4 standard chest X-ray segmentation datasets: JSRT \cite{jsrt_shiraishi2000development} (246 subjects), Padchest \cite{bustos2020padchest} (137 subjects), Montgomery \cite{montgomeryset} (138 subjects), and Shenzhen \cite{shenzhen} (390 subjects). The dataset is publicly available at \url{https://github.com/ngaggion/Chest-xray-landmark-dataset}.

The multi-center composition creates a heterogeneous label scenario identical to that addressed in \cite{gaggion_isbi}, where JSRT provides comprehensive annotations for lungs, heart, and clavicles, Padchest includes annotations for lungs and heart, while Montgomery and Shenzhen contain only lung annotations. This distribution results in 246 subjects with complete annotations, 383 subjects total with both lung and heart annotations available, and all 911 subjects having lung annotations. Each dataset maintains its original imaging characteristics and acquisition protocols, preserving the domain shift challenges that make this a realistic multi-center scenario. Following \cite{gaggion_isbi}, we use the same 80\% train/validation and 20\% test partitions for each dataset, yielding 197/49 subjects for JSRT, 110/27 for Padchest, 111/27 for Montgomery, and 312/78 for Shenzhen in the train/test splits respectively.

We evaluate four architectural configurations exploring two key design dimensions: decoder architecture (single vs. dual) and the adjancency matrix construction strategy (independent vs. unified). Figure~\ref{fig:organ_contours} illustrates these representation strategies. Independent graph representations model each anatomical structure separately with circular graphs (left lung in blue, right lung in green, heart in orange, left clavicle in cyan, right clavicle in purple), while unified graph representations enable modeling of inter-organ spatial relationships by allowing anatomically adjacent structures to share connection nodes (shown in cyan). All models incorporate differentiable rasterization for complementary pixel-level supervision alongside Chamfer distance and edge-based regularization losses (see \ref{app:rasterization} for ablation study on the effect of rasterization).

Figure~\ref{fig:correspondences} shows the learned landmark correspondence patterns for both independent and unified graph representations. The visualizations reveal that our models develop consistent anatomical correspondences across the population, where corresponding landmark positions represent similar anatomical locations across different subjects.

\begin{table*}[t!]
\centering
\caption{Performance comparison on CAMUS dataset showing Dice coefficients (DC), Hausdorff distances (HD), and Average Symmetric Surface Distances (ASSD) for cardiac structure segmentation.}
\label{tab:camus_results}
\resizebox{\textwidth}{!}{
\begin{tabular}{ll ccc ccc ccc}
\toprule
& & \multicolumn{3}{c}{\textbf{LV Endo}} & \multicolumn{3}{c}{\textbf{LV Epi}} & \multicolumn{3}{c}{\textbf{LA}} \\ 
\cmidrule(lr){3-5} \cmidrule(lr){6-8} \cmidrule(lr){9-11}
\textbf{Model} & \textbf{Graph Type} & \textbf{DC $\uparrow$} & \textbf{HD $\downarrow$} & \textbf{ASSD $\downarrow$} & \textbf{DC $\uparrow$} & \textbf{HD $\downarrow$} & \textbf{ASSD $\downarrow$} & \textbf{DC $\uparrow$} & \textbf{HD $\downarrow$} & \textbf{ASSD $\downarrow$} \\ 
\midrule
nnUNet & - & 0.940 ± 0.027 & 13.686 ± 5.890 & 4.575 ± 1.807 & 0.873 ± 0.048 & 17.203 ± 6.698 & 4.889 ± 1.740 & 0.900 ± 0.076 & 16.782 ± 12.305 & 5.491 ± 4.048 \\ 
\midrule
Mask-HybridGNet & Independent & 0.925 ± 0.042 & 17.285 ± 7.151 & 5.819 ± 3.021 & 0.840 ± 0.076 & 22.181 ± 8.755 & 6.405 ± 2.557 & 0.884 ± 0.083 & 19.105 ± 11.609 & 6.563 ± 4.837 \\ 
Mask-HybridGNet Dual & Independent & 0.929 ± 0.039 & 16.665 ± 7.997 & 5.452 ± 2.764 & 0.850 ± 0.065 & 21.279 ± 8.182 & 5.951 ± 2.213 & 0.889 ± 0.074 & 17.939 ± 11.155 & 6.192 ± 4.065 \\ 
Mask-HybridGNet & Unified & 0.926 ± 0.038 & 18.678 ± 8.240 & 5.699 ± 2.625 & 0.850 ± 0.058 & 21.713 ± 9.529 & 5.896 ± 2.262 & 0.875 ± 0.094 & 18.984 ± 10.305 & 6.836 ± 4.638 \\ 
Mask-HybridGNet Dual & Unified & \textbf{0.930 ± 0.035} & \textbf{17.296 ± 7.777} & \textbf{5.382 ± 2.453} & \textbf{0.855 ± 0.061} & \textbf{20.182 ± 8.487} & \textbf{5.645 ± 2.184} & \textbf{0.886 ± 0.085} & \textbf{18.234 ± 11.393} & \textbf{6.223 ± 4.133} \\ 
\bottomrule
\end{tabular}}
\end{table*}

Table~\ref{tab:chestxraycomparison} presents quantitative results comparing our Mask-HybridGNet variants against the landmark-supervised baseline and state-of-the-art nnUNet. Our framework achieves performance comparable to the baseline across all datasets and anatomical structures, successfully validating that graph-based models can be trained with mask supervision without sacrificing accuracy. The dual decoder architecture consistently outperforms single decoder variants, demonstrating the benefits of enhanced feature sharing between convolutional and graph pathways, where the graph decoder leverages boundary-focused representations that have already been optimized through pixel-level supervision. Unified graph representations show modest but consistent improvements over independent representations, particularly for organs where modeling shared boundaries provides anatomical context. Even though they are not directly comparable, we include the state-of-the-art nnUNet model as a reference. While it achieves slightly superior raw performance metrics, our approach provides capabilities unavailable to pixel-based methods: guaranteed topological consistency and implicit anatomical correspondence learning enabling population-level analysis.

\begin{figure*}[!ht]
    \centering
    \includegraphics[width=\linewidth]{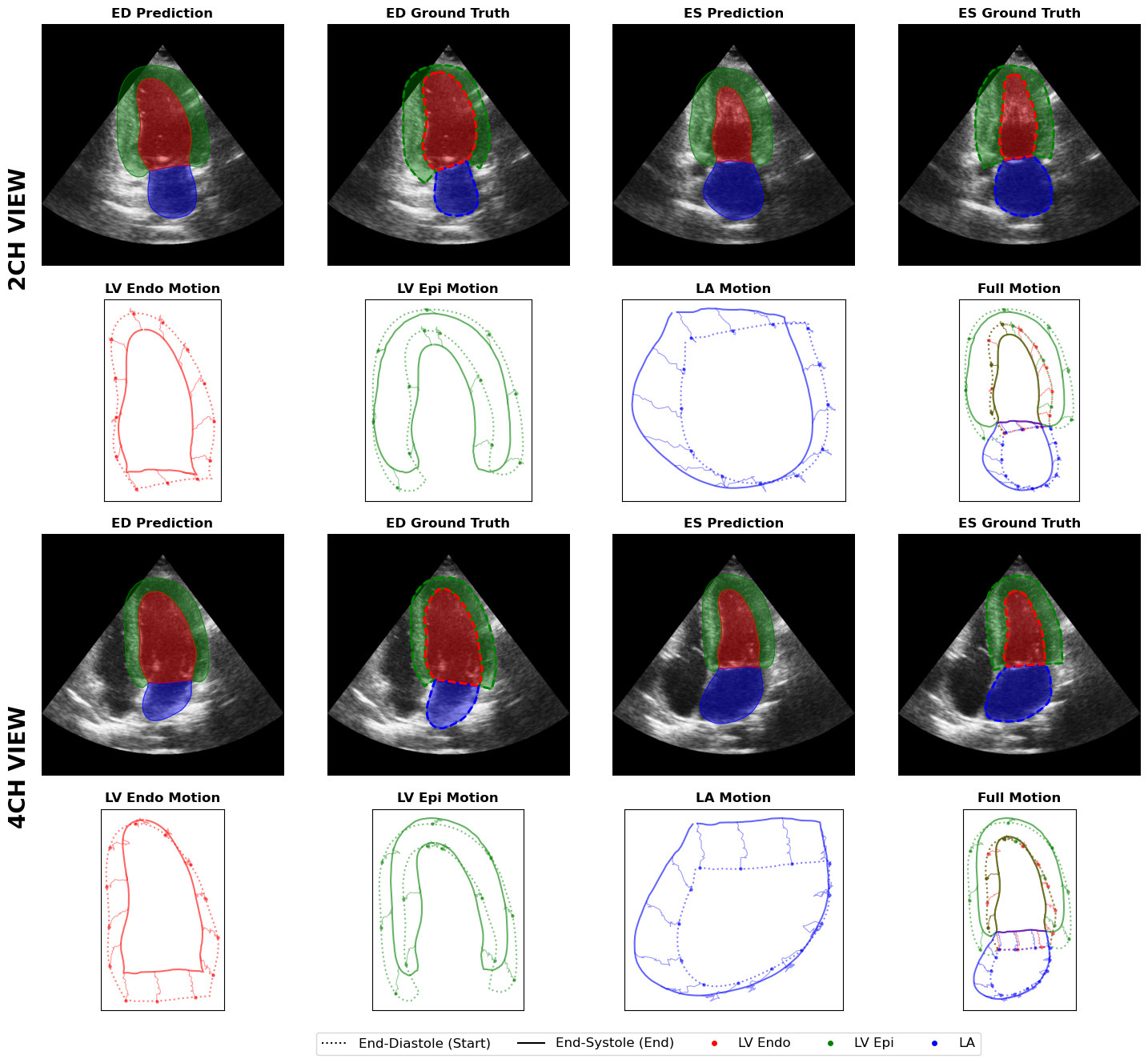}
    \caption{\textbf{Temporal cardiac analysis demonstrating landmark correspondence tracking using a unified graph representation.} The figure displays comprehensive cardiac segmentation and tracking for a representative test patient with normal systolic function (EF = 54\%) across two-chamber (top two rows) and four-chamber (bottom two rows) views. For each view, the first row compares the model prediction against the ground truth at end-diastole and end-systole. The second row visualizes the temporal motion tracking of landmarks for the left ventricular endocardium (LV Endo), epicardium (LV Epi), and left atrium (LA).}
    \label{fig:cardiac_cycle}
\end{figure*}

\subsection{Evaluating Temporal Landmark Consistency and Atlas Creation on Ultrasound Heart Image Segmentation}

For our second experiment, we evaluated our framework on the CAMUS (Cardiac Acquisitions for Multi-structure Ultrasound Segmentation) dataset \cite{CAMUS}, which comprises cardiac ultrasound exams from 500 patients acquired at the University Hospital of St Etienne, France, covering multiple cardiac pathologies.

The dataset exhibits significant clinical heterogeneity that reflects real-world echocardiographic imaging challenges. The patient population includes 35\% good quality images, 46\% medium quality images, and 19\% poor quality images, with half of the patients presenting left ventricular ejection fraction (LVEF) below 45\%, indicating pathological risk. Each patient's data includes 2D apical four-chamber and two-chamber view sequences acquired using GE Vivid E95 ultrasound scanners with standardized protocols. The dataset provides manual expert annotations for the left ventricular endocardium (LV Endo), the left ventricular epicardium (LV Epi), and the left atrium (LA),  performed by experienced cardiologists.

We evaluated both single decoder (Mask-HybridGNet) and dual decoder (Mask-HybridGNet Dual) architectures with independent and unified graph representations, as illustrated in Figure \ref{fig:organ_contours}. All models use three resolution levels for hierarchical graph representation (see \ref{app:resolution_levels} for ablation studies for resolution level selection).

Table~\ref{tab:camus_results} presents the evaluation results comparing our Mask-HybridGNet variants against nnUNet across three cardiac structures. Our graph-based approach achieves competitive performance while providing inherent topological guarantees. Mask-HybridGNet Dual with unified graph representations achieves the best performance among our variants. Figure \ref{fig:cardiac_cycle} demonstrates a key advantage of our landmark-based representation: generating one-to-one correspondences across the cardiac cycle enables precise tracking of cardiac motion patterns and morphological changes throughout the complete temporal sequence for both imaging views.

For temporal tracking applications, the unified contours representation demonstrates significant advantages over independent contour processing. The unified approach explicitly models shared boundaries between cardiac structures, maintaining topological consistency where the endocardium and epicardium walls meet. This produces anatomically coherent tracking results that preserve proper spatial relationships between cardiac structures throughout the cardiac cycle. In contrast, the independent contours approach processes each structure separately, which can lead to boundary discontinuities at organ interfaces where separate contours fail to maintain consistent shared boundaries, resulting in anatomically implausible gaps or overlaps between adjacent structures. Figure \ref{fig:cardiac_cycle_independent} in  \ref{app:independent_camus_contours} illustrates these boundary discontinuity issues.

\subsubsection{Anatomical Correspondences Generation from Automated Segmentation Models}

So far, we have experimented with generating contour-based segmentations with correspondences directly from images. However, we may also be interested in extracting structured anatomical correspondences from existing segmentation masks. This raises a critical practical question: can models trained to extract anatomical correspondences from \textit{ground truth segmentations} operate effectively on automated segmentation outputs? This would enable the generation of anatomically corresponding landmarks from any high-quality segmentation model without requiring specialized retraining.

For this task, we utilize the Mask-HybridGNet architecture (Figure~\ref{fig:architecture}, left). Since the input is a pre-computed segmentation mask rather than a raw image, the model works as a shape auto-encoder: the variational encoder-decoder maps the dense binary input to a structured latent representation, which is then decoded into the final graph. The auxiliary pixel-wise branch becomes redundant in this configuration. We trained two separate models using ground truth segmentation masks as input: one with a unified graph representation and another with an independent graph representation.

To assess whether the anatomical correspondence learning process is robust to different segmentation sources, these models were evaluated on two test sets consisting of subjects not present in the training partition of either model: the manual GT masks and automated nnUNet-generated segmentations. The nnUNet outputs were post-processed to retain only the largest connected component. This was necessary to ensure topological validity, as the graph-based model is inherently incapable of representing the spurious, disconnected pixel clusters often produced by pixel-wise classifiers, and we will measure the performance by evaluating the model capacity to reconstruct the input in a graph representation.

Table~\ref{tab:reconstruction_fidelity} presents the agreement between the generated graph segmentation and the input mask, on the complete CAMUS test set ($n=1932$ images). The results demonstrate remarkable robustness. The model achieves virtually identical performance regardless of the input source, with the ASSD remaining consistently below 1 pixel for all structures and graph types. This sub-pixel level of precision indicates that the graph decoder captures the geometric details of the input masks with high fidelity while enforcing necessary smoothness.

\begin{table*}[t!]
\centering
\caption{Agreement between graph segmentation outputs and input masks. Metrics evaluate Dice, HD and ASSD between the output of the Mask-HybridGNet model and the specific input mask (GT vs. nnUNet) provided.}
\label{tab:reconstruction_fidelity}
\begin{tabular}{ll ccc ccc ccc}
\toprule
& & \multicolumn{2}{c}{\textbf{Dice Coefficient $\uparrow$}} & \multicolumn{2}{c}{\textbf{HD (pixels) $\downarrow$}} & \multicolumn{2}{c}{\textbf{ASSD (pixels) $\downarrow$}} \\
\cmidrule(lr){3-4} \cmidrule(lr){5-6} \cmidrule(lr){7-8}
\textbf{Organ} & \textbf{Graph Type} & \textbf{GT Input} & \textbf{nnUNet Input} & \textbf{GT Input} & \textbf{nnUNet Input} & \textbf{GT Input} & \textbf{nnUNet Input} \\
\midrule
LA & Independent & 0.988 ± 0.005 & 0.986 ± 0.004 & 3.06 ± 1.425 & 2.834 ± 1.216 & 0.709 ± 0.204 & 0.746 ± 0.131 \\
LA & Unified & 0.99 ± 0.004 & 0.989 ± 0.003 & 2.852 ± 1.053 & 2.532 ± 0.707 & 0.613 ± 0.183 & 0.595 ± 0.084 \\
\midrule
LV Endo & Independent & 0.99 ± 0.002 & 0.989 ± 0.002 & 3.096 ± 0.978 & 2.774 ± 0.694 & 0.769 ± 0.133 & 0.748 ± 0.087 \\
LV Endo & Unified & 0.991 ± 0.002 & 0.99 ± 0.002 & 6.224 ± 3.407 & 4.472 ± 1.866 & 0.7 ± 0.102 & 0.683 ± 0.063 \\
\midrule
LV Epi & Independent & 0.979 ± 0.005 & 0.981 ± 0.004 & 5.181 ± 2.622 & 3.663 ± 1.871 & 0.842 ± 0.152 & 0.734 ± 0.09 \\
LV Epi & Unified & 0.983 ± 0.003 & 0.984 ± 0.003 & 3.281 ± 1.372 & 2.612 ± 0.684 & 0.678 ± 0.101 & 0.619 ± 0.041 \\
\bottomrule
\end{tabular}
\end{table*}

Notably, the reconstruction error (measured by HD and ASSD) is often lower for nnUNet inputs than for the manual ground truth. This is because manual segmentations frequently contain noise, such as jagged edges and non-smooth transitions resulting from the manual tracing process, which can be seen in Figure \ref{fig:cardiac_cycle}. Conversely, the automated nnUNet predictions, once stripped of spurious artifacts, seems to provide a smoother boundary that aligns more naturally with the geometric regularity enforced by the graph decoder.

This consistency validates that the learned anatomical correspondence patterns are stable across different segmentation sources. For applications requiring pixel-perfect alignment, these landmarks can be post-processed to snap exactly to the segmentation boundary. This maintains the learned correspondences while achieving absolute agreement with the input, combining the benefits of structured atlas-based analysis with the raw accuracy of state-of-the-art pixel-based models.

Consequently, institutions could maintain their preferred automated segmentation pipelines while utilizing our framework as a shape-regularizing post-processor to gain anatomical correspondence, enabling downstream morphological analysis and temporal tracking without architectural changes.

\subsection{Cardiac MRI Segmentation: Sunnybrook Cardiac Data Evaluation}

To further validate our framework's applicability across different cardiac imaging modalities, we conducted an experiment on the Sunnybrook Cardiac Data (SCD), a well-established benchmark dataset from the 2009 Cardiac MR Left Ventricle Segmentation Challenge. The SCD comprises 45 cine-MRI images acquired as 2D 256$\times$256 images, with each slice treated as an independent 2D image for segmentation. The dataset focuses specifically on left ventricular structures, providing expert annotations for the left ventricular endocardium (LV Endo) and epicardium (LV Epi), from which the myocardium region (LV Myo) can be derived as the space between them. It represents a diverse pathological spectrum including healthy subjects, hypertrophy cases, heart failure with infarction, and heart failure without infarction.

In this experiment, we evaluated both Mask-HybridGNet and Mask-HybridGNet Dual using independent graph representations for the LV Endo and LV Epi with three resolution levels. Unlike the multi-chamber views in CAMUS, these structures are nested; since there are no shared interfaces or contact walls requiring joint modeling, independent circular graphs provide the most straightforward and topologically appropriate representation. Furthermore, we leveraged this dataset to perform an ablation study on graph resolution levels, the results of which are detailed in \ref{app:resolution_levels}, justifying the selection of the multi-resolution hierarchy.

\begin{figure}[t!]
\centering
\includegraphics[width=\linewidth]{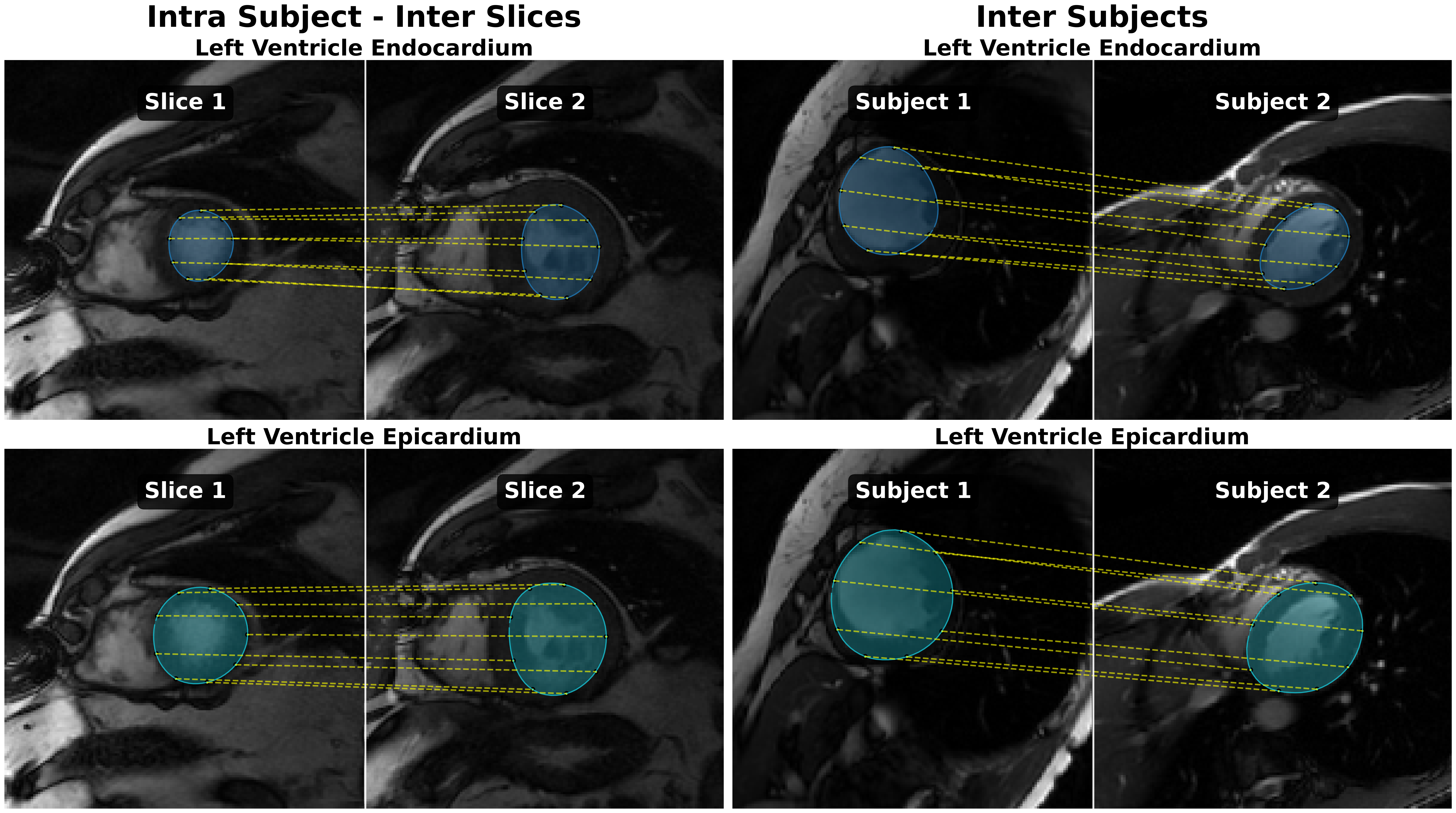}
\caption{\textbf{Learned landmark correspondences across cardiac MRI slices demonstrate both cross-slice intra-subject and inter-subject tracking capabilities.} The figure shows two different slices from the same subject in the Sunnybrook Cardiac Data (left), two different slices from different subjects (right).}
\label{fig:sunnybrook_correspondences}
\end{figure}

\begin{table}[t!]
\centering
\caption{Performance comparison on Sunnybrook Cardiac Data showing Dice coefficients (DC), Hausdorff distances (HD), and Average Symmetric Surface Distances (ASSD) for left ventricular segmentation.}
\label{tab:sunnybrook_results}
\resizebox{\linewidth}{!}{
\begin{tabular}{ll ccc ccc}
\toprule
& & \multicolumn{3}{c}{\textbf{LV Endo}} & \multicolumn{3}{c}{\textbf{LV Myo}} \\ 
\cmidrule(lr){3-5} \cmidrule(lr){6-8}
\textbf{Model} & \textbf{Graph Type} & \textbf{DC $\uparrow$} & \textbf{HD $\downarrow$} & \textbf{ASSD $\downarrow$} & \textbf{DC $\uparrow$} & \textbf{HD $\downarrow$} & \textbf{ASSD $\downarrow$} \\ 
\midrule
nnUNet & - & 0.919 & 3.559 & 1.186 & 0.734 & 3.979 & 0.997 \\ 
\midrule
Mask-HybridGNet & Independent & 0.893 & 4.014 & 1.535 & 0.655 & 4.508 & 1.385 \\ 
Mask-HybridGNet Dual & Independent & \textbf{0.903} & \textbf{3.744} & \textbf{1.417} & \textbf{0.673} & \textbf{4.553} & \textbf{1.305} \\ 
\bottomrule
\end{tabular}}
\end{table}

Table~\ref{tab:sunnybrook_results} presents the performance metrics for both model variants alongside an nnUNet baseline. While the pixel-based approach achieves higher Dice scores—particularly in the myocardium where the thin, ring-like geometry is highly sensitive to landmark placement—our graph-based models provide distinct structural advantages. Mask-HybridGNet Dual achieves a competitive performance for the endocardium, and both variants ensure anatomically plausible, closed boundaries that are inherently free from the fragmentation common in pixel-wise classifiers.

\begin{table*}[!t]
\centering
\caption{Multi-dataset fetal head and pubic symphysis segmentation results. Metrics (Dice, HD, ASSD) are reported for training on individual datasets (HC18, JNU-IFM, PSFHS) versus a combined training set (Multi-centric). Models were evaluated across all test sets to assess cross-dataset generalization. We compare our Mask-HybridGNet Dual (independent contours) against nnUNet. $N$ indicates the number of test images for each evaluation set. N/A indicates that no ground truth data for pubic symphisis is available in this dataset.}
\label{tab:fetal_head_multi_dataset}
\resizebox{\textwidth}{!}{
\begin{tabular}{l ccc c ccc c ccc c}
\toprule
& \multicolumn{4}{c}{\textbf{Tested on HC18}} & \multicolumn{4}{c}{\textbf{Tested on JNU-IFM}} & \multicolumn{4}{c}{\textbf{Tested on PSFHS}} \\
\cmidrule(lr){2-5} \cmidrule(lr){6-9} \cmidrule(lr){10-13}
\textbf{Model} & \textbf{DC $\uparrow$} & \textbf{HD $\downarrow$} & \textbf{ASSD $\downarrow$} & \textbf{N} & \textbf{DC $\uparrow$} & \textbf{HD $\downarrow$} & \textbf{ASSD $\downarrow$} & \textbf{N} & \textbf{DC $\uparrow$} & \textbf{HD $\downarrow$} & \textbf{ASSD $\downarrow$} & \textbf{N} \\
\midrule
\multicolumn{13}{c}{\textbf{Fetal Head Segmentation}} \\
\midrule
\textit{Mask-HybridGNet Dual} & & & & & & & & & & & & \\
Multi-centric & 0.970 & 14.942 & 5.440 & 101 & \textbf{0.951} & \textbf{31.664} & \textbf{11.695} & 406 & \textbf{0.924} & \textbf{12.302} & \textbf{4.393} & 510 \\
HC18-only & \textbf{0.975} & \textbf{11.375} & \textbf{4.416} & 101 & 0.827 & 97.057 & 40.554 & 406 & 0.802 & 28.611 & 11.154 & 510 \\
JNU-IFM-only & 0.884 & 50.873 & 21.854 & 101 & 0.931 & 43.841 & 16.642 & 406 & 0.884 & 17.408 & 7.051 & 510 \\
PSFHS-only & 0.893 & 47.447 & 19.754 & 101 & 0.947 & 33.952 & 12.462 & 406 & 0.913 & 13.383 & 5.046 & 510 \\
\textit{nnUNet} & & & & & & & & & & & & \\
Multi-centric & 0.976 & 14.196 & 2.964 & 101 & 0.356 & 52.423 & 21.641 & 406 & 0.804 & 69.823 & 16.293 & 510 \\
HC18-only & \textbf{0.978} & \textbf{7.098} & \textbf{2.521} & 101 & 0.760 & 67.911 & 18.423 & 406 & 0.810 & 74.533 & 21.664 & 510 \\
JNU-IFM-only & 0.952 & 23.149 & 6.188 & 101 & 0.356 & 60.911 & 21.089 & 406 & 0.710 & 144.637 & 29.869 & 510 \\
PSFHS-only & 0.930 & 36.740 & 8.947 & 101 & 0.810 & 70.611 & 19.211 & 406 & 0.872 & 79.389 & 17.683 & 510 \\
\midrule
\multicolumn{13}{c}{\textbf{Pubic Symphysis Segmentation}} \\
\midrule
\textit{Mask-HybridGNet Dual} & & & & & & & & & & & & \\
Multi-centric & \multicolumn{4}{c}{N/A} & \textbf{0.893} & \textbf{27.153} & \textbf{7.751} & 312 & \textbf{0.823} & \textbf{10.318} & \textbf{3.527} & 510 \\
JNU-IFM-only & \multicolumn{4}{c}{N/A} & 0.848 & 37.093 & 10.955 & 312 & 0.353 & 76.842 & 55.679 & 510 \\
PSFHS-only & \multicolumn{4}{c}{N/A} & 0.876 & 30.969 & 9.054 & 312 & 0.834 & 9.415 & 3.228 & 510 \\
\textit{nnUNet} & & & & & & & & & & & & \\
Multi-centric & \multicolumn{4}{c}{N/A} & 0.892 & 9.112 & 2.395 & 312 & 0.877 & 15.571 & 4.466 & 510 \\
JNU-IFM-only & \multicolumn{4}{c}{N/A} & 0.786 & 14.907 & 4.431 & 312 & 0.338 & 19.539 & 8.781 & 510 \\
PSFHS-only & \multicolumn{4}{c}{N/A} & 0.929 & 8.384 & 1.970 & 312 & 0.873 & 15.805 & 4.471 & 510 \\
\bottomrule
\end{tabular}}
\end{table*}

Beyond segmentation accuracy, the graph-based representation captures stable anatomical reference points across the cardiac volume. As shown in Figure~\ref{fig:sunnybrook_correspondences}, the learned landmark correspondences remain consistent across different slices of the same subject. While the model also shows some level of emergent correspondence across different subjects that could be further investigated, the 2D nature of the input slices means that intra-subject consistency provides the most reliable basis for clinical application. This ability to maintain consistent anatomical relationships within a subject's volume may enable applications such as temporal cardiac motion tracking within cine-MRI sequences or cross-slice reconstruction for comprehensive 3D cardiac analysis without the need for additional registration steps.

\begin{figure*}[t]
\centering
\includegraphics[width=\linewidth]{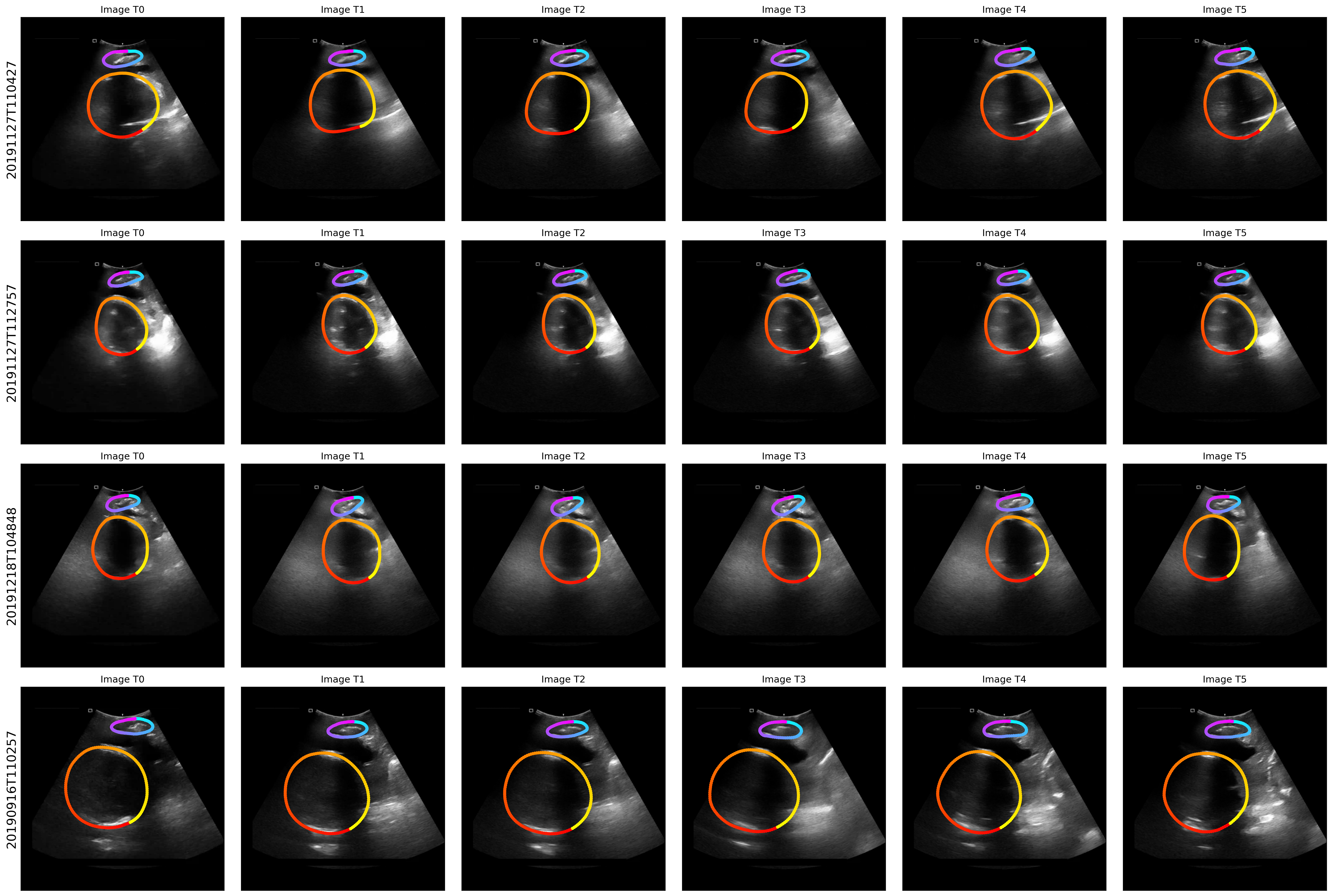}
\caption{\textbf{Temporal landmark correspondence demonstration across multiple patients.} Color-coded landmark visualization showing consistent anatomical correspondences for fetal head (autumn colormap) and pubic symphysis (cool colormap) structures across four representative patients from the JNU-IFM dataset. Each row represents a different patient's temporal sequence, demonstrating the learned anatomical atlas where corresponding landmark indices represent the same anatomical locations across the population. This capability enables clinical applications including standardized measurement, temporal pattern analysis, and population-based shape studies.}
\label{fig:landmark_correspondence_fetal}
\end{figure*}

\subsection{Evaluating the Robustness on Multi-centric Data in Fetal Head Segmentation}

We conducted a multi-centric evaluation using three publicly available fetal ultrasound datasets with distinct imaging protocols and clinical applications. This experiment demonstrates our framework's ability to learn from heterogeneous data sources while maintaining anatomical correspondence across different patient populations and acquisition settings.

Fetal head segmentation in ultrasound presents well-documented challenges. The comprehensive Challenge US study \cite{rueda2013evaluation} identified several key difficulties: (i) image quality varies dramatically within clinical datasets; (ii) segmentation difficulty increases substantially with gestational age due to progressive bone mineralization causing acoustic shadowing; and (iii) the appearance of anatomical structures shows high variability due to fetal position, motion artifacts, and operator-dependent acquisition.

We utilized three datasets representing distinct clinical contexts: the HC18 dataset~\cite{hc18paper} containing 999 two-dimensional ultrasound images for fetal head circumference measurement across all trimesters; the JNU-IFM dataset~\cite{jnu-ifm} comprising 5203 intrapartum transperineal ultrasound images with both fetal head and mother's pubic symphysis annotations from 78 videos across 51 patients; and the PSFHS dataset~\cite{psfhs} featuring 5101 annotated intrapartum images from multiple clinical centers. HC18 focuses on prenatal biometry, while JNU-IFM and PSFHS address intrapartum monitoring for labor progression assessment.

For this evaluation, we employed Mask-HybridGNet Dual with an independent graph representations. We also trained nnUNet models for comparison using the same training configurations. For each dataset, we performed subject-wise random splitting with 10\% of subjects allocated to testing, ensuring no data leakage between training and evaluation phases. We evaluated four training configurations for both methods: individual dataset training (HC18-only, JNU-IFM-only, PSFHS-only) and the multi-centric dataset training combining all three datasets. Each trained model was evaluated across all available test sets to assess both within-dataset performance and cross-dataset generalization capabilities.

Figure~\ref{fig:landmark_correspondence_fetal} demonstrates the temporal landmark correspondence capabilities enabled by our approach using representative sequences from the JNU-IFM dataset. The visualization shows four patients with color-coded landmark positions for both fetal head and pubic symphysis structures, revealing consistent anatomical correspondences across video frames despite variations in fetal head size, orientation, and imaging conditions. 

Table~\ref{tab:fetal_head_multi_dataset} presents the evaluation results across all datasets and training configurations. The multi-centric dataset training consistently demonstrates superior cross-dataset generalization compared to individual dataset training. For fetal head segmentation, the multi-centric model achieves balanced performance across all test sets with Dice coefficients of 0.970, 0.951, and 0.924 for HC18, JNU-IFM, and PSFHS respectively. Models trained on individual datasets show substantial performance degradation when evaluated on different datasets, highlighting domain shift challenges across institutions and protocols. For pubic symphysis segmentation, multi-centric training shows even more pronounced benefits, achieving Dice coefficients of 0.893 and 0.823 for JNU-IFM and PSFHS test sets. The dramatic performance difference between JNU-IFM-trained and multi-centric models on PSFHS data (DC: 0.353 vs 0.823) underscores the critical importance of diverse training data for robust multi-institutional deployment.

In contrast to our method's robust generalization, nnUNet demonstrates significant sensitivity to annotation heterogeneity in this multi-centric scenario. Both nnUNet multi-centric and nnUNet JNU-IFM achieve only 0.356 DC on the JNU-IFM test set for fetal head segmentation. This failure comes from a known limitation of pixel-based methods when trained with heterogeneous annotations \cite{gaggion_isbi}: when structures are inconsistently annotated across training samples, these models learn unpredictable patterns about when to produce segmentations. Consequently, the model produces empty predictions for some images while segmenting others, even within the same dataset and imaging conditions, making its behavior unreliable for clinical deployment. For pubic symphysis, nnUNet-JNU-IFM similarly fails on PSFHS (0.338 DC vs our 0.823 DC). Our graph-based approach avoids this failure mode by guaranteeing closed contour predictions by construction, ensuring anatomically plausible segmentations are always generated regardless of annotation inconsistencies during training.

\begin{figure*}[ht!]
    \centering
    \includegraphics[width=\linewidth]{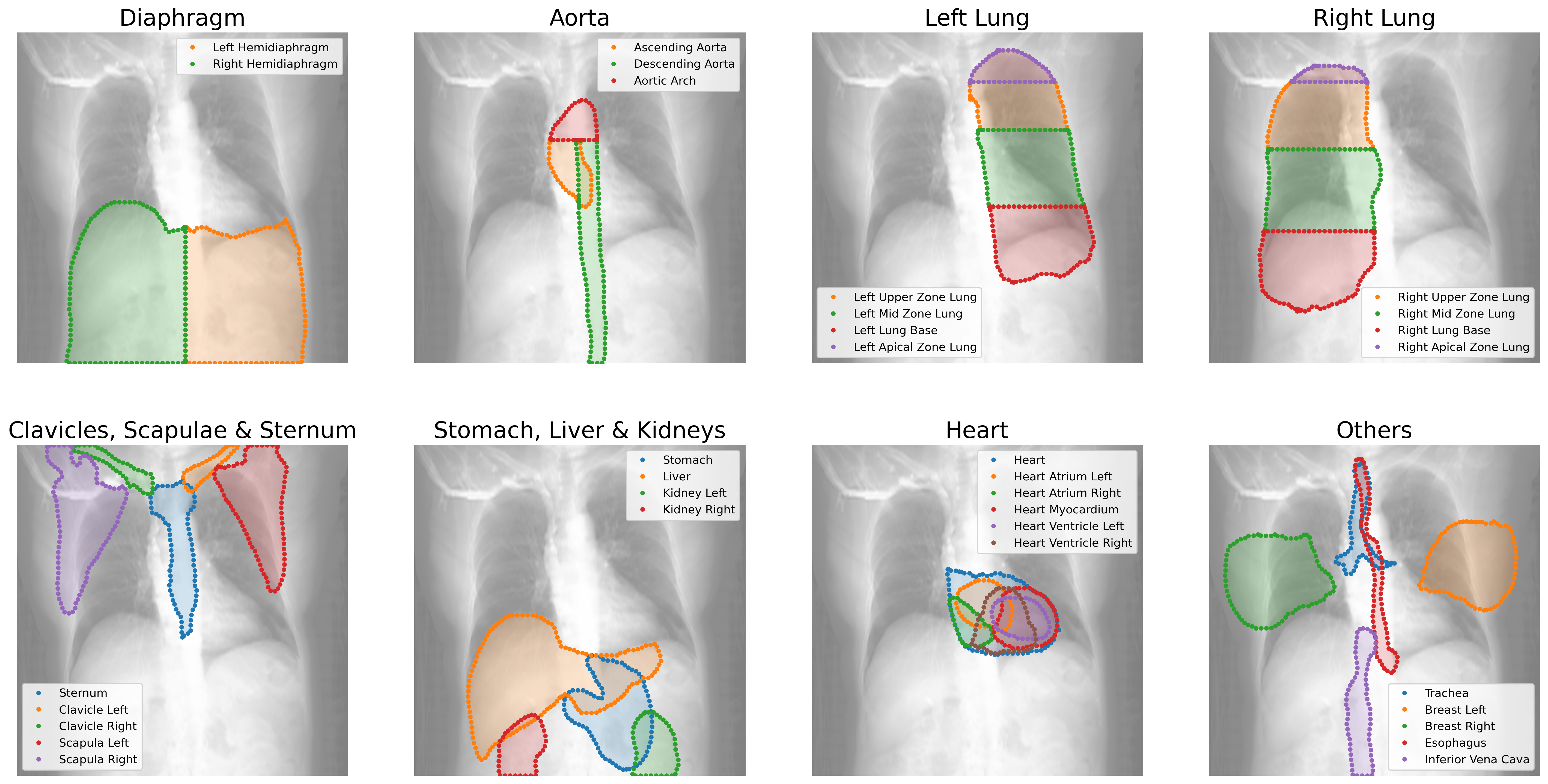}
    \caption{\textbf{Anatomical atlas and organ grouping strategy for PAX-Ray++ chest radiograph segmentation.} The figure displays a representative chest X-ray atlas with colored anatomical structures organized into eight functional categories. Top row shows unified graph representations: (1) Diaphragm Complex including diaphragm, left and right hemidiaphragms; (2) Aortic System comprising aorta, ascending aorta, descending aorta, and aortic arch; (3) Left Lung Zones encompassing left lung and its four anatomical zones; (4) Right Lung Zones including right lung and its four anatomical zones. Bottom row presents independent contour groupings: (5) Clavicles, Scapulae \& Sternum; (6) Stomach, Liver \& Kidneys; (7) Heart structures including chambers and myocardium; (8) Other soft tissues comprising trachea, breast tissue, esophagus, and inferior vena cava.}
    \label{fig:paxray_organs}
\end{figure*}

\begin{figure*}[ht!]
    \centering
    \includegraphics[width=\linewidth]{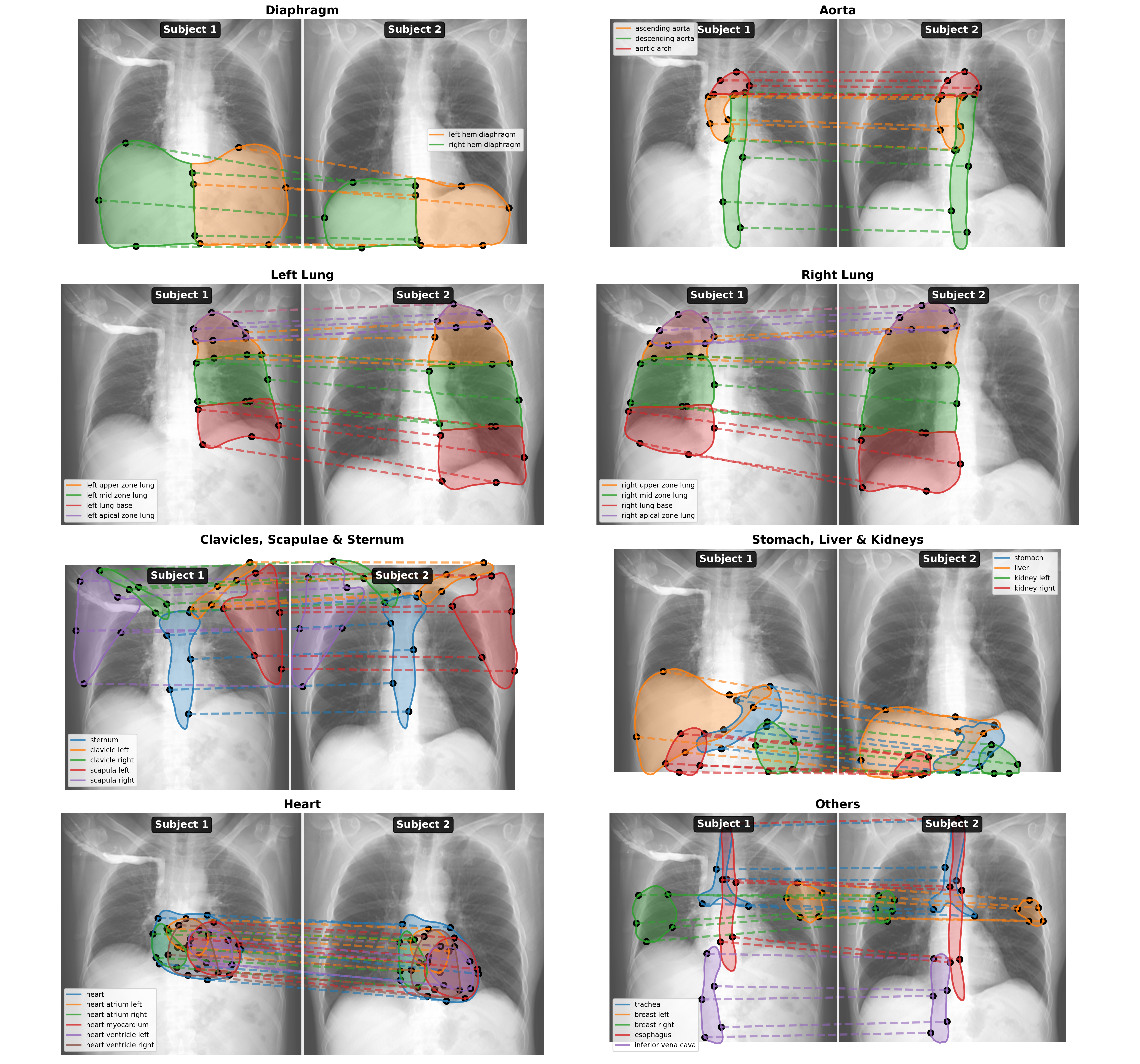}
    \caption{\textbf{Emergent anatomical correspondences learned through mask supervision on PAX-Ray++ dataset.} The figure demonstrates that our framework successfully learns anatomically meaningful landmark correspondences without explicit point-level supervision across two representative chest X-ray subjects. Each row shows a different organ group with Subject 1 (left) and Subject 2 (right) displayed side-by-side. Yellow dashed lines connect corresponding landmarks between subjects, revealing that landmark indices represent consistent anatomical locations across the population.}
    \label{fig:paxray_correspondence}
\end{figure*}

\subsection{Large-Scale Chest Radiograph Anatomy Segmentation: PAX-Ray++ Dataset Application}

\begin{table}[!ht]
\centering
\caption{Performance comparison on PAX-Ray++ dataset showing Dice coefficients (DC), Hausdorff distances (HD), and Average Symmetric Surface Distances (ASSD) for 37-organ chest radiograph segmentation across 737 test images.}
\label{tab:paxray_results}
\resizebox{\linewidth}{!}{
\begin{tabular}{ll ccc}
\toprule
\textbf{Model} & \textbf{Graph Type} & \textbf{DC $\uparrow$} & \textbf{HD $\downarrow$} & \textbf{ASSD $\downarrow$} \\
\midrule
Mask-HybridGNet & Independent & 0.837 ± 0.142 & 25.269 ± 20.081 & 7.391 ± 9.170 \\
Mask-HybridGNet & Unified & 0.855 ± 0.134 & 23.185 ± 19.914 & 6.617 ± 9.236 \\
Mask-HybridGNet Dual & Independent & 0.864 ± 0.133 & 22.894 ± 19.777 & 6.299 ± 9.063 \\
Mask-HybridGNet Dual & Unified & \textbf{0.869 ± 0.132} & \textbf{22.321 ± 20.192} & \textbf{6.148 ± 9.409} \\
\bottomrule
\end{tabular}}
\end{table}

To demonstrate that our model can scale to multi-organ segmentation scenarios, we applied it to the PAX-Ray++ dataset \cite{paxray}, moving beyond the typical 3-5 organ structures examined in previous studies to a clinically comprehensive 37-organ chest radiograph analysis. PAX-Ray++ provides pixel-wise segmentation masks for 157 anatomical structures across 14,754 chest X-ray images derived from projected CT scans using volumetric pseudo-labeling techniques.

For this experiment, we selected 37 anatomically meaningful structures that form well-defined contours suitable for graph-based representation, focusing exclusively on frontal view projections where clinical measurements are typically performed. Due to this selection, we worked with 7,377 images, using half of the dataset for this experiment. The selected organs span all major anatomical systems visible in chest radiographs: cardiovascular (11 structures including heart chambers, myocardium, and major vessels), respiratory (11 structures encompassing lungs, lung zones, and trachea), diaphragmatic (3 structures including diaphragm and hemidiaphragms), abdominal (4 structures including liver, kidneys, and stomach), musculoskeletal (5 structures comprising sternum, clavicles, and scapulae), and other clinically relevant soft tissues (3 structures including breast tissue and esophagus).

We implemented a hybrid approach combining unified and independent graph representations based on anatomical relationships and shared boundary characteristics. Figure \ref{fig:paxray_organs} illustrates our organ grouping strategy across eight categories. We identified four anatomical systems that benefit from joint modeling due to shared boundaries and functional relationships: the Diaphragm Complex (diaphragm, left hemidiaphragm, right hemidiaphragm), the Aortic System (aorta, ascending aorta, descending aorta, aortic arch), Left Lung Zones (left lung, left upper zone lung, left mid zone lung, left lung base, left apical zone lung), and Right Lung Zones (right lung, right upper zone lung, right mid zone lung, right lung base, right apical zone lung). These unified representations capture the natural anatomical continuity within organ systems while enabling detailed subregional analysis. The remaining 21 organs maintain independent circular graph representations, grouped conceptually for visualization as: Clavicles, Scapulae \& Sternum (5 organs), Stomach, Liver \& Kidneys (4 organs), Heart structures (6 organs), and Other soft tissues (6 organs).

For unified graph construction, we employed a single representative chest X-ray as the anatomical atlas to define spatial relationships and shared boundaries between organs within each group. The unified contour generation process, explained in \ref{app:unified_construction}, successfully identified interface regions where adjacent structures share anatomical boundaries, particularly effective for the naturally continuous structures like the diaphragm complex and segmented aortic system. 

We evaluated all four architectural configurations with both independent and unified graph representations. Table \ref{tab:paxray_results} presents the evaluation results. Mask-HybridGNet Dual with unified graph representations achieves the best overall performance. Figure \ref{fig:paxray_correspondence} demonstrates the learned landmark correspondences between two representative subjects across all organ groups, with yellow correspondence lines connecting equivalent anatomical locations between subjects.

This large-scale application demonstrates that our contour-based approach scales effectively to comprehensive chest radiograph analysis, successfully handling 37 anatomical structures while preserving topological guarantees (detailed per-organ analysis in Appendix Table~\ref{tab:paxray_detailed}).

\section{Discussion}

This work enables graph-based segmentation from pixel-based mask annotations eliminating its fundamental barrier: the requirement for manually annotated landmarks with point-to-point correspondences. Beyond this practical contribution, our experiments reveal an emergent property: anatomical correspondence arises naturally from optimization without explicit supervision. The architectural explanation is straightforward: our model outputs a fixed number of nodes arranged through a fixed adjacency matrix. For the graph decoder's learned features to generalize across patients, each node index must consistently represent similar anatomical locations—otherwise, if node 15 represented "apex" in one patient and "lateral wall" in another, the features learned for position 15 would be meaningless. This architectural constraint - combined with Chamfer distance supervision which establishes soft matches through nearest-neighbor associations and edge-based regularization enforcing connectivity - drives landmarks to settle into geometrically stable configurations. Visualizations across chest radiography, cardiac ultrasound, cardiac MRI, and fetal imaging datasets demonstrate that corresponding landmark indices in the predicted boundary graphs represent consistent anatomical locations across the population. This connects our approach to decades of statistical shape modeling research, where correspondences have traditionally relied on geometric saliency or manual definition, and reveals that consistent correspondence can instead emerge from as a consequence of the model inductive bias, rather than explicit supervision, when the graph structure imposes consistent topology. Such consistency enables robust and objective estimation of clinically meaningful biomarkers from 2D segmentations anchored to anatomically grounded landmarks. 

The distinction between our approach and pixel-based methods becomes particularly evident in our multi-centric fetal experiments. When trained on heterogeneous annotations from multiple institutions, nnUNet exhibited catastrophic failure \cite{gaggion_isbi}, producing empty predictions for some images while segmenting others within the same dataset and imaging conditions. 
Our graph-based approach avoids this failure mode by guaranteeing closed contour predictions by construction, ensuring anatomically plausible segmentations are always generated regardless of annotation inconsistencies during training. 

The minor differences in performance relative to nnUNet reflects different optimization objectives rather than capability limitations. nnUNet's architecture is tuned for pixel-wise accuracy through ensemble strategies, test-time augmentation, and sliding window inference. Our framework optimizes for segmentation accuracy under topological and correspondence constraints—a fundamentally different objective function. This represents a tradeoff between pure accuracy and geometric interpretability: while nnUNet excels at pixel-level overlap metrics, it cannot provide topological guarantees or anatomical correspondences. Our segmentation-to-landmarks experiments on CAMUS demonstrate these approaches are not mutually exclusive: we successfully extracted anatomical atlases from nnUNet-generated segmentations, enabling hybrid deployment strategies that combine nnUNet's accuracy with our correspondence capabilities. This addresses a practical reality: research groups with existing high-performing segmentation pipelines can augment them with correspondence extraction without retraining. 

Current limitations include the restriction to 2D applications, though this reflects the challenge of extending mask-supervised training to 3D rather than a fundamental constraint. While HybridVNet \cite{gaggion2025multi} demonstrates that graph-based methods can model volumetric structures as surface meshes, adapting our Chamfer-based training strategy to match variable-length 3D surface contours extracted from volumetric masks against fixed-size mesh predictions remains an open problem. The framework is also limited to structures with simple topology (closed boundaries) and cannot directly model anatomically complex structures with branching (e.g., vascular trees) or holes without fundamental extensions to the graph representation. Finally, careful adjacency matrix design is required when modeling unified graphs to correctly encode which structures share boundaries, and computational cost scales with the number of organs due to rasterization loss backpropagation overhead.


Future work should explore extending the framework to 3D volumetric segmentation and investigating whether correspondence quality can be further improved through additional self-supervised losses, such as cycle consistency across temporal sequences. Furthermore, future iterations could incorporate a priori defined anatomical landmarks, such as the cardiac apex or valve insertions, to be explicitly mapped into the graph representation. This would transition the framework from discovering purely emergent correspondences to a constrained atlas-based model that aligns with specific clinical measurement protocols.

By removing the landmark annotation requirement while preserving graph-based segmentation advantages—topological guaranties, anatomical correspondence, shape priors— this work makes structured anatomical modeling accessible to existing medical imaging datasets. The finding that correspondences emerge from architectural necessity rather than explicit supervision may inspire investigation into other emergent properties of geometrically-constrained learning. Our public release of trained models, code, and correspondence extraction tools aims to lower barriers for researchers exploring atlas-based analysis, temporal tracking, and shape-constrained segmentation as complementary capabilities alongside pixel-based methods.

\section*{Acknowledgments}
EF was supported by the Google Award for Inclusion Research, a Googler Initiated Grant and the Invited Professor program of the DataIA Paris-Saclay Institute. EF and MV are supported by the STIC-AmSud CGFLRVE project. MJL acknowledge the support of Spanish Ministerio de Ciencia e Innovación, Agencia Estatal de Investigación, under grant PID2022-141493OB-I00 (10.13039/501100011033/MCIN/AEI/ERDF, UE), co-financed by European Regional Development Fund (ERDF), ‘A way of making Europe’ and partial funding from the MAGERIT-CM project (TEC-2024/COM-44), supported by the Research and Development Activities Program by the Comunidad de Madrid. MV is also supported by the ANR-21-CE45-0007 and the ANR-23-IACL-0003 – DATAIA CLUSTER (as part of IA CLUSTER program).

\section*{Declaration of generative AI and AI-assisted technologies in the manuscript preparation process}

During the preparation of this work the authors used Gemini (Google) in order to improve the grammar and readability of the manuscript. After using this service, the authors reviewed and edited the content as needed and take full responsibility for the content of the published article. All scientific results, interpretations, and citations were generated and verified by the authors.

\bibliographystyle{plain}
\bibliography{bib.bib}

\clearpage

\appendix

\section{Dataset Processing Procedures}
\label{app:config}

\subsection{Data Organization and Format Requirements}

Medical imaging datasets exhibit significant heterogeneity in data organization and storage formats. Input images may be stored as PNG files, DICOM series, time-series sequences, or institution-specific formats, while segmentation masks can exist as PNG files, NIfTI volumes (.nii.gz), or distributed across separate annotation files. Due to this variability, our framework requires users to implement custom data reading functions for both images and masks that understand their specific dataset structure.

The framework implements subject-wise data splitting following each dataset's original patient organization structure, provided the custom reading functions properly identify patient associations. This ensures all images from the same patient remain within the same partition (training, validation, or testing), preventing information leakage between splits—a critical consideration for medical imaging applications.

\subsection{Configuration System}

For each dataset, we define a configuration dictionary that encapsulates all processing parameters:

\begin{verbatim}
CONFIG = {
    "database_path": "../Source/dataset",
    "output_path": "../Output/dataset", 
    "scale_factor": 0.10,
    "resolutions": ["Full", "Half", "Quarter"],
    "organs": ["1", "2"],
    "organ_names": ["Name 1", "Name 2"],
    "inputsize": 512,
    "flip_h": True,
    "flip_v": False,
    "rotate": True,
    "transpose": False
}
\end{verbatim}

Each parameter serves a specific function in the processing pipeline. The \texttt{database\_path} and \texttt{output\_path} parameters specify source and destination directories, allowing the framework to work with arbitrary file structures while maintaining organized output hierarchies. The \texttt{scale\_factor} controls the trade-off between boundary detail preservation and computational efficiency by determining what proportion of the average contour length to use for fixed-size landmark representations—typically set between 0.05 and 0.15 depending on anatomical complexity.

The \texttt{resolutions} parameter defines the multi-resolution hierarchy through string identifiers, with the framework automatically generating appropriate downsampling and upsampling operators based on the specified levels. The \texttt{organs} parameter lists numerical class identifiers present in the segmentation masks, while \texttt{organ\_names} provides corresponding descriptive labels for human-readable output and visualization. The \texttt{inputsize} parameter standardizes processing resolution for the neural network while preserving original image resolution during preprocessing.

Data augmentation parameters control geometric transformations applied during training. The \texttt{flip\_h} and \texttt{flip\_v} parameters enable horizontal and vertical flipping respectively, \texttt{rotate} allows rotational augmentation within specified angular ranges, and \texttt{transpose} permits matrix transposition operations. These boolean flags are configured per-dataset to respect anatomical constraints.

\subsection{Contour Extraction and Preprocessing}

The contour extraction process begins by creating binary masks for each anatomical structure of interest. For multi-class segmentation masks, we extract individual anatomical structures by selecting specific class values, while for datasets with separate organ channels, we apply appropriate thresholding operations. This organ-independent processing ensures each extracted contour represents a single connected anatomical component.

Boundary contours are extracted using external contour retrieval methods with no contour approximation to preserve complete boundary detail and maintain the original coordinate precision. Each extracted contour maintains its original variable length and is stored alongside corresponding organ identifiers. The resulting ground truth representations consist of ordered sequences of 2D coordinate points that trace anatomical boundaries in their native resolution and spatial relationship.

\subsection{Graph Connectivity Preprocessing}

To enable efficient computation of edge-based regularization losses during training, we precompute and organize edge connectivity information during the dataset organization phase. This preprocessing transforms graph connectivity information into vectorized representations that support batch-wise parallel computation across all organs and edges simultaneously.

For each resolution level, we extract three essential data structures from the adjacency matrices: (i) a matrix containing all edge connections for each organ, (ii) a mapping that identifies which organ each edge belongs to, and (iii) the adjacency relationships between consecutive edges in each organ's contour.

These precomputed structures eliminate the need for iterative edge traversal during training, significantly improving computational efficiency for the edge-based contour regularization terms. The vectorized representations enable simultaneous computation of uniform edge length, elasticity, and curvature losses across the entire batch, regardless of the number of organs or their individual sizes.

\subsection{Dataset Adaptation Framework}

Adapting the framework to new medical imaging datasets involves two essential steps: creating a configuration dictionary that specifies processing parameters, and implementing dataset-specific functions to handle image and mask loading.

We provide comprehensive Jupyter notebook examples for each dataset used in this study. These notebooks demonstrate the complete adaptation workflow, including configuration setup, custom data reading function implementation, and dataset-specific pre-processing steps. Each notebook serves as a template that can be modified for new datasets with similar characteristics.

Once the configuration dictionary and data reading functions are established, the processing pipeline automatically generates appropriate graph representations, adjacency matrices, and multi-resolution hierarchies based on the extracted contour statistics. This approach provides a systematic framework for incorporating new datasets while ensuring users understand their data characteristics and can implement appropriate quality control measures.

\section{Unified Graph Representation Construction Details}
\label{app:unified_construction}

\subsection{Atlas Definition and Organ Grouping}

Unlike independent organ contours defined using dataset statistics (average contour lengths), unified representations require an atlas mask, a representative multi-label segmentation that defines spatial relationships between anatomical structures. This atlas serves as the template for extracting contour coordinates and identifying shared organ boundaries.

The atlas mask can be obtained through several approaches: (1) selecting a representative subject from the training set that exhibits typical anatomical relationships, (2) using an externally generated anatomical atlas, or (3) creating a synthetic mask that represents the most generic anatomical configuration possible for the target structures.

For instance, in pulmonary imaging, we would define two atlas masks: one containing the left lung subdivisions (lower lobe, upper lobe) and another containing the right lung subdivisions (lower lobe, middle lobe, upper lobe). Each atlas mask provides the spatial template needed to identify shared boundaries between adjacent regions.

\subsection{Contour Extraction and Boundary Identification}

Unlike independent organ representations that rely on dataset statistics, unified representations require extracting actual contours from the atlas mask to identify where organs physically contact each other. For each atlas mask, we extract individual contours $\mathbf{C}^{(o)} = \{\mathbf{c}_1^{(o)}, \mathbf{c}_2^{(o)}, ..., \mathbf{c}_{L^{(o)}}^{(o)}\}$ for each labeled region $o \in \mathcal{O}$ from the atlas mask using OpenCV contour extraction methods.

We then implement a proximity-based merging algorithm that identifies points lying within a specified distance threshold $\delta$ of each other across different organ contours. Points satisfying the proximity criterion are merged into single nodes representing shared boundary regions. Each merged node maintains a membership set indicating which organs it belongs to, enabling the model to recognize and leverage inter-organ relationships during processing.

The threshold parameter $\delta$ is typically set based on the image resolution and expected boundary precision. In practice, we set $\delta = \sqrt{2} + \epsilon$ where $\epsilon$ is a small positive constant (e.g., $\epsilon = 0.01$) that ensures diagonally adjacent pixels are captured while avoiding numerical precision issues at exactly $\delta = \sqrt{2}$.

\section{Topologically-Aware Multi-Resolution Processing}
\label{app:topological_processing}

\subsection{Topologically-Aware Contour Downsampling}

Our downsampling approach preserves topological structure by distinguishing between simple contour segments and junction points. We define node degree as $\deg(v) = |\{u \in V : A_{v,u} = 1\}|$. Nodes with $\deg(v) = 2$ qualify for downsampling, while nodes with $\deg(v) \geq 3$ are junction points and are always preserved across all resolution levels $r$.

For downsampling, we identify adjacent pairs $(u, v)$ connected by an edge ($A_{u,v} = 1$) where both nodes are candidates. These pairs are merged, with positions and organ memberships computed as:

\begin{align}
\mathbf{p}'_{(u,v)} &= \frac{\mathbf{p}_u + \mathbf{p}_v}{2} \\
\mathcal{M}'_{(u,v)} &= \mathcal{M}_u \cup \mathcal{M}_v
\end{align}

where $\mathbf{p}_u$ and $\mathbf{p}_v$ are the positions of nodes $u$ and $v$, and $\mathcal{M}_u$ and $\mathcal{M}_v$ are their respective organ membership sets.

\subsection{Downsampling and Upsampling Matrices}

Unlike independent organ representations that use separate matrices for each organ, unified graphs require only single downsampling and upsampling matrices that operate on the entire graph structure. For each resolution level $r \in \{1, 2, \ldots, R\}$, we generate downsampling matrices $\mathbf{D}^{r \rightarrow r+1} \in \mathbb{R}^{|V_{r+1}| \times |V_r|}$ and upsampling matrices $\mathbf{U}^{r+1 \rightarrow r} \in \mathbb{R}^{|V_r| \times |V_{r+1}|}$, where $V_r$ and $V_{r+1}$ denote the node sets at resolutions $r$ and $r+1$ respectively, that encode the relationship between consecutive resolution levels.

The downsampling matrix elements are defined as:

\begin{equation}
\mathbf{D}^{r \rightarrow r+1}_{i,j} = 
\begin{cases}
1 & \text{if node } j \in V_r \text{ is preserved as node } i \in V_{r+1} \\
0.5 & \text{if node } j \text{ is averaged to create } i \\
0 & \text{otherwise}
\end{cases}
\end{equation}

Similarly, the upsampling matrix elements are defined as:

\begin{equation}
\mathbf{U}^{r+1 \rightarrow r}_{i,j} = 
\begin{cases}
1 & \text{if node } i \in V_r \text{ corresponds to node } j \in V_{r+1} \\
0 & \text{otherwise}
\end{cases}
\end{equation}

Finally, the adjacency structure at resolution $r+1$ inherits connectivity from resolution $r$: edges connect nodes whose constituent nodes were connected at the finer level, naturally preserving contour topology.

\begin{table*}[tp]
\centering
\caption{Complete ablation study showing the effect of differentiable rasterization on all architectural variants on the Chest-xray-landmark dataset in terms of Dice score.}
\label{tab:rasterization_ablation}
\resizebox{\linewidth}{!}{
\begin{tabular}{lll cccc cc c}
\toprule
& & & \multicolumn{4}{c}{\textbf{Lungs}} & \multicolumn{2}{c}{\textbf{Heart}} & \textbf{Clavicles} \\ 
\cmidrule(lr){4-7} \cmidrule(lr){8-9} \cmidrule(lr){10-10}
\textbf{Architecture} & \textbf{Graph} & \textbf{Rast.} & \textbf{Shenzhen} & \textbf{Montgomery} & \textbf{Padchest} & \textbf{JSRT} & \textbf{Padchest} & \textbf{JSRT} & \textbf{JSRT} \\ 
\midrule
Single Decoder & Independent & No  & 0.958 ± 0.014 & 0.964 ± 0.021 & 0.946 ± 0.020 & 0.966 ± 0.012 & 0.935 ± 0.029 & 0.935 ± 0.032 & 0.825 ± 0.078 \\
               &             & Yes & 0.961 ± 0.015 & 0.968 ± 0.018 & 0.953 ± 0.016 & 0.967 ± 0.014 & 0.925 ± 0.038 & 0.928 ± 0.038 & 0.801 ± 0.067 \\ 
\midrule
Dual Decoder   & Independent & No  & 0.961 ± 0.014 & 0.969 ± 0.019 & 0.950 ± 0.020 & 0.970 ± 0.011 & 0.930 ± 0.027 & 0.936 ± 0.032 & 0.841 ± 0.062 \\
               &             & Yes & 0.965 ± 0.014 & 0.974 ± 0.019 & 0.956 ± 0.022 & 0.975 ± 0.011 & 0.939 ± 0.019 & 0.941 ± 0.027 & 0.860 ± 0.077 \\ 
\midrule
Single Decoder & Unified     & No  & 0.960 ± 0.014 & 0.966 ± 0.020 & 0.948 ± 0.019 & 0.968 ± 0.012 & 0.931 ± 0.028 & 0.936 ± 0.033 & 0.827 ± 0.059 \\
               &             & Yes & 0.963 ± 0.014 & 0.974 ± 0.015 & 0.954 ± 0.017 & 0.972 ± 0.011 & 0.940 ± 0.022 & 0.935 ± 0.034 & 0.836 ± 0.068 \\ 
\midrule
Dual Decoder   & Unified     & No  & 0.962 ± 0.014 & 0.972 ± 0.018 & 0.951 ± 0.018 & 0.971 ± 0.012 & 0.939 ± 0.020 & 0.939 ± 0.029 & 0.854 ± 0.099 \\
               &             & Yes & 0.965 ± 0.013 & 0.975 ± 0.018 & 0.955 ± 0.020 & 0.975 ± 0.011 & 0.942 ± 0.017 & 0.940 ± 0.027 & 0.862 ± 0.078 \\ 
\bottomrule
\end{tabular}}
\end{table*}

\section{Ablation Study on Differentiable Rasterization}
\label{app:rasterization}

To evaluate the contribution of differentiable rasterization, we trained all architectural variants both with and without this component. Table~\ref{tab:rasterization_ablation} shows that incorporating rasterization consistently improves performance across all configurations, increasing Dice scores across different organs. This validates our design choice to include rasterization as a standard component of Mask-HybridGNet, as it provides complementary pixel-level supervision that enhances the Chamfer distance-based contour matching.

\section{Ablation Study on Contour Resolution Levels}
\label{app:resolution_levels}

We conducted a comprehensive evaluation to determine the optimal number of resolution levels for our hierarchical graph representation across different cardiac imaging datasets. Our framework employs multi-resolution graph representations where each resolution level captures anatomical structures at different scales, from global shape context to fine boundary details. We systematically evaluated architectures using 2, 3, and 4 resolution levels to identify optimal configurations, investigating the trade-offs between segmentation accuracy, computational efficiency, and training time.

\subsection{CAMUS Dataset Resolution Analysis}

Table~\ref{tab:camus_resolution} presents the performance comparison across different resolution levels on the CAMUS echocardiography dataset. Statistical analysis using Wilcoxon tests reveals distinct behavior patterns between architectures:

For Mask-HybridGNet, performance remains stable between 2 and 3 resolution levels (p = 0.52), but degrades significantly at 4 resolutions (2 vs 4: p $<$ 0.001; 3 vs 4: p $<$ 0.001). This indicates that adding more resolution levels beyond 3 provides no benefit and actually harms performance.

For Mask-HybridGNet Dual, performance improves significantly from 2 to 3 resolution levels (p $<$ 1e-11) and from 2 to 4 resolutions (p $<$ 1e-14). However, there is no significant difference between 3 and 4 resolutions (p = 0.15), indicating that the additional computational cost of 4 resolutions does not provide meaningful performance gains.

\begin{table}[h!]
\centering
\caption{Performance comparison on CAMUS dataset showing Dice coefficients (DC), Hausdorff distances (HD), and Average Symmetric Surface Distances (ASSD) across different resolution levels.}
\label{tab:camus_resolution}
\resizebox{\linewidth}{!}{
\begin{tabular}{ll ccc ccc}
\toprule
& & \multicolumn{3}{c}{\textbf{Mask-HybridGNet}} & \multicolumn{3}{c}{\textbf{Mask-HybridGNet Dual}} \\
\cmidrule(lr){3-5} \cmidrule(lr){6-8}
\textbf{Structure} & \textbf{Res.} & \textbf{DC $\uparrow$} & \textbf{HD $\downarrow$} & \textbf{ASSD $\downarrow$} & \textbf{DC $\uparrow$} & \textbf{HD $\downarrow$} & \textbf{ASSD $\downarrow$} \\
\midrule
LV Myo  & 2 & 0.925 & 17.589 & 5.819 & 0.924 & 17.299 & 5.871 \\
        & 3 & \textbf{0.926} & \textbf{17.679} & \textbf{5.738} & \textbf{0.931} & \textbf{15.768} & \textbf{5.309} \\
        & 4 & 0.921 & 18.403 & 6.099 & 0.933 & 15.911 & 5.118 \\
\midrule
LV Endo & 2 & 0.841 & 23.248 & 6.182 & 0.843 & 21.457 & 5.968 \\
        & 3 & \textbf{0.836} & \textbf{22.724} & \textbf{6.216} & \textbf{0.849} & \textbf{20.839} & \textbf{5.711} \\
        & 4 & 0.833 & 24.960 & 6.446 & 0.854 & 20.768 & 5.526 \\
\midrule
LA      & 2 & 0.873 & 19.881 & 6.923 & 0.877 & 19.362 & 6.797 \\
        & 3 & \textbf{0.878} & \textbf{19.873} & \textbf{6.670} & \textbf{0.890} & \textbf{17.941} & \textbf{6.036} \\
        & 4 & 0.871 & 20.324 & 7.005 & 0.889 & 17.386 & 6.008 \\
\bottomrule
\end{tabular}
}
\end{table}

\subsection{Sunnybrook Dataset Resolution Analysis}

Table~\ref{tab:sunnybrook_resolution} presents the resolution level evaluation on the Sunnybrook cardiac MRI dataset. Statistical analysis reveals markedly different behavior patterns compared to CAMUS:

For Mask-HybridGNet, performance improves significantly with increasing resolution levels (3 vs 2: p $<$ 0.001; 4 vs 2: p $<$ 1e-11; 4 vs 3: p = 0.001).

For Mask-HybridGNet Dual, performance remains stable between 2 and 3 resolution levels (p = 0.90), with no significant difference between 3 and 4 resolutions (p = 0.14), but shows slight degradation at 4 compared to 2 resolutions (p = 0.03).

\begin{table}[h!]
\centering
\caption{Performance comparison on Sunnybrook Cardiac Data showing Dice coefficients (DC), Hausdorff distances (HD), and Average Symmetric Surface Distances (ASSD) across different resolution levels.}
\label{tab:sunnybrook_resolution}
\resizebox{\linewidth}{!}{
\begin{tabular}{ll ccc ccc}
\toprule
& & \multicolumn{3}{c}{\textbf{Mask-HybridGNet}} & \multicolumn{3}{c}{\textbf{Mask-HybridGNet Dual}} \\
\cmidrule(lr){3-5} \cmidrule(lr){6-8}
\textbf{Structure} & \textbf{Res.} & \textbf{DC $\uparrow$} & \textbf{HD $\downarrow$} & \textbf{ASSD $\downarrow$} & \textbf{DC $\uparrow$} & \textbf{HD $\downarrow$} & \textbf{ASSD $\downarrow$} \\
\midrule
LV Endo & 2 & 0.882 & 4.302 & 1.690 & 0.904 & 3.654 & 1.396 \\
        & 3 & 0.893 & 4.014 & 1.535 & \textbf{0.903} & \textbf{3.744} & \textbf{1.417} \\
        & 4 & \textbf{0.900} & \textbf{3.664} & \textbf{1.416} & 0.887 & 3.887 & 1.601 \\
\midrule
LV Myo  & 2 & 0.598 & 5.557 & 1.589 & 0.684 & 4.434 & 1.295 \\
        & 3 & 0.655 & 4.508 & 1.385 & \textbf{0.673} & \textbf{4.553} & \textbf{1.305} \\
        & 4 & \textbf{0.669} & \textbf{4.484} & \textbf{1.282} & 0.659 & 4.707 & 1.463 \\
\bottomrule
\end{tabular}
}
\end{table}

\subsection{Summary}

The resolution level analysis reveals that 3 resolution levels provide an effective balance between segmentation accuracy and computational efficiency across both cardiac imaging modalities. While Mask-HybridGNet could benefit from 4 resolution levels on certain datasets, the marginal improvements do not justify the increased training time. Mask-HybridGNet Dual demonstrates robust performance at 3 resolution levels across all datasets, making this configuration our standard choice for all experiments.

\clearpage

\begin{onecolumn}

\section{On the usage of independent graph representations}
\label{app:independent_camus_contours}

\begin{figure*}[h!]
    \centering
    \includegraphics[width=\linewidth]{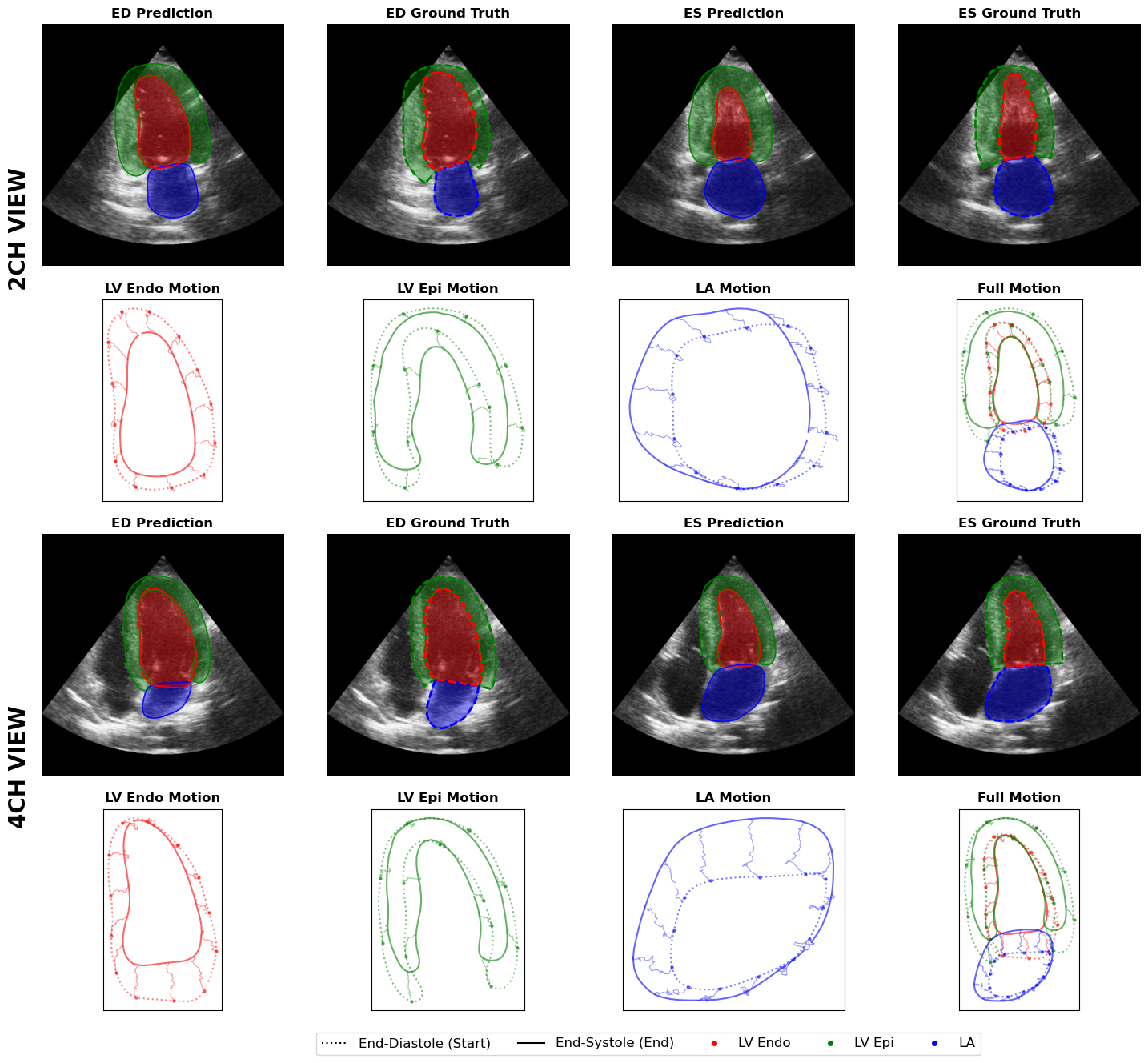}
    \caption{\textbf{Temporal cardiac analysis demonstrating landmark correspondence tracking using independent graphs representation.} The figure displays comprehensive cardiac segmentation and tracking for a representative test patient  with normal systolic function (EF = 54\%) across two-chamber (top two rows) and four-chamber (bottom two rows) views. For each view, the first row compares the model prediction against the ground truth at end-diastole and end-systole. The second row visualizes the temporal motion tracking of landmarks for the left ventricular endocardium (LV Endo), epicardium (LV Epi), and left atrium (LA). Unlike the unified approach, this independent representation fails to enforce shared boundaries, resulting in anatomically implausible overlaps and gaps at the interface between the left atrium and left ventricle.}
    \label{fig:cardiac_cycle_independent}
\end{figure*}

\clearpage

\section{Detailed PAX-Ray++ Performance on individual organs}

\begin{table*}[!ht]
\centering
\caption{Detailed per-organ performance comparison on PAX-Ray++ dataset. The table presents Dice coefficients (mean ± standard deviation) for all 37 anatomical structures across four model configurations. Results demonstrate the consistent benefits of the dual decoder architecture and unified graph representations across diverse anatomical systems.}
\label{tab:paxray_detailed}
\begin{tabular}{l cc cc}
\toprule
& \multicolumn{2}{c}{\textbf{Mask-HybridGNet}} & \multicolumn{2}{c}{\textbf{Mask-HybridGNet Dual}} \\
\cmidrule(lr){2-3} \cmidrule(lr){4-5}
\textbf{Structure} & \textbf{Independent} & \textbf{Unified} & \textbf{Independent} & \textbf{Unified} \\
\midrule
Diaphragm & 0.961 ± 0.027 & 0.967 ± 0.027 & 0.945 ± 0.029 & 0.966 ± 0.024 \\
Left Hemidiaphragm & 0.947 ± 0.044 & 0.946 ± 0.044 & 0.923 ± 0.049 & 0.943 ± 0.045 \\
Right Hemidiaphragm & 0.964 ± 0.023 & 0.967 ± 0.025 & 0.950 ± 0.027 & 0.965 ± 0.023 \\
\midrule
Aorta & 0.886 ± 0.072 & 0.897 ± 0.067 & 0.867 ± 0.070 & 0.883 ± 0.068 \\
Ascending Aorta & 0.856 ± 0.090 & 0.852 ± 0.091 & 0.831 ± 0.097 & 0.841 ± 0.093 \\
Descending Aorta & 0.852 ± 0.097 & 0.858 ± 0.089 & 0.822 ± 0.096 & 0.842 ± 0.091 \\
Aortic Arch & 0.840 ± 0.098 & 0.830 ± 0.106 & 0.791 ± 0.125 & 0.814 ± 0.103 \\
\midrule
Left Lung & 0.957 ± 0.043 & 0.958 ± 0.045 & 0.944 ± 0.045 & 0.954 ± 0.043 \\
Left Upper Zone Lung & 0.934 ± 0.056 & 0.934 ± 0.062 & 0.915 ± 0.062 & 0.928 ± 0.056 \\
Left Mid Zone Lung & 0.910 ± 0.070 & 0.915 ± 0.075 & 0.899 ± 0.072 & 0.911 ± 0.069 \\
Left Lung Base & 0.880 ± 0.096 & 0.885 ± 0.095 & 0.854 ± 0.106 & 0.880 ± 0.093 \\
Left Apical Zone Lung & 0.859 ± 0.123 & 0.852 ± 0.130 & 0.815 ± 0.130 & 0.844 ± 0.118 \\
\midrule
Right Lung & 0.959 ± 0.028 & 0.959 ± 0.030 & 0.952 ± 0.027 & 0.953 ± 0.025 \\
Right Upper Zone Lung & 0.929 ± 0.045 & 0.930 ± 0.049 & 0.920 ± 0.048 & 0.923 ± 0.049 \\
Right Mid Zone Lung & 0.910 ± 0.067 & 0.910 ± 0.069 & 0.904 ± 0.067 & 0.908 ± 0.072 \\
Right Lung Base & 0.876 ± 0.130 & 0.879 ± 0.128 & 0.862 ± 0.126 & 0.870 ± 0.129 \\
Right Apical Zone Lung & 0.840 ± 0.125 & 0.852 ± 0.125 & 0.812 ± 0.135 & 0.832 ± 0.122 \\
\midrule
Sternum & 0.831 ± 0.071 & 0.838 ± 0.067 & 0.814 ± 0.081 & 0.830 ± 0.067 \\
Clavicle Left & 0.820 ± 0.123 & 0.841 ± 0.109 & 0.738 ± 0.145 & 0.781 ± 0.117 \\
Clavicle Right & 0.825 ± 0.123 & 0.841 ± 0.110 & 0.758 ± 0.129 & 0.790 ± 0.112 \\
Scapula Left & 0.881 ± 0.069 & 0.895 ± 0.067 & 0.859 ± 0.067 & 0.867 ± 0.061 \\
Scapula Right & 0.887 ± 0.067 & 0.892 ± 0.068 & 0.854 ± 0.065 & 0.868 ± 0.064 \\
\midrule
Stomach & 0.752 ± 0.126 & 0.761 ± 0.128 & 0.719 ± 0.127 & 0.748 ± 0.123 \\
Liver & 0.920 ± 0.053 & 0.925 ± 0.053 & 0.902 ± 0.056 & 0.919 ± 0.053 \\
Kidney Left & 0.674 ± 0.266 & 0.673 ± 0.271 & 0.599 ± 0.260 & 0.660 ± 0.257 \\
Kidney Right & 0.605 ± 0.292 & 0.604 ± 0.299 & 0.534 ± 0.274 & 0.582 ± 0.284 \\
\midrule
Heart & 0.947 ± 0.031 & 0.949 ± 0.031 & 0.934 ± 0.035 & 0.944 ± 0.029 \\
Heart Atrium Left & 0.881 ± 0.070 & 0.886 ± 0.067 & 0.864 ± 0.078 & 0.879 ± 0.069 \\
Heart Atrium Right & 0.890 ± 0.073 & 0.893 ± 0.068 & 0.870 ± 0.078 & 0.882 ± 0.068 \\
Heart Myocardium & 0.916 ± 0.057 & 0.917 ± 0.056 & 0.897 ± 0.062 & 0.913 ± 0.054 \\
Heart Ventricle Left & 0.893 ± 0.075 & 0.897 ± 0.072 & 0.874 ± 0.082 & 0.891 ± 0.074 \\
Heart Ventricle Right & 0.884 ± 0.074 & 0.884 ± 0.071 & 0.863 ± 0.078 & 0.883 ± 0.068 \\
\midrule
Inferior Vena Cava & 0.774 ± 0.151 & 0.775 ± 0.145 & 0.752 ± 0.148 & 0.766 ± 0.143 \\
Esophagus & 0.741 ± 0.118 & 0.769 ± 0.115 & 0.719 ± 0.116 & 0.742 ± 0.118 \\
Trachea & 0.819 ± 0.071 & 0.845 ± 0.067 & 0.781 ± 0.076 & 0.795 ± 0.079 \\
Breast Left & 0.689 ± 0.214 & 0.695 ± 0.206 & 0.675 ± 0.207 & 0.693 ± 0.205 \\
Breast Right & 0.714 ± 0.211 & 0.708 ± 0.213 & 0.689 ± 0.206 & 0.704 ± 0.216 \\
\bottomrule
\end{tabular}
\end{table*}

\end{onecolumn}

\end{document}